\documentclass[twoside,11pt]{article}
\usepackage{jmlr2e}
\usepackage{bm}
\usepackage{amsmath}
\usepackage{amssymb,amsfonts}
\usepackage{algorithm}
\usepackage[noend]{algpseudocode}
\algrenewcommand\algorithmicthen{}
\algrenewcommand\algorithmicdo{}

\usepackage{graphicx}
\usepackage{xcolor}

\usepackage[style=apa,natbib=true,doi=false,url=false]{biblatex}
\usepackage[per-mode=symbol]{siunitx}
\usepackage[capitalise]{cleveref}
\usepackage[section]{placeins}
\usepackage{soul}

\usepackage{listings}
\usepackage{subcaption}
\usepackage{pgfplots}
\pgfplotsset{compat=newest}
\pgfplotsset{every axis legend/.append style={%
cells={anchor=west}}
}
\definecolor{lead_color}{RGB}{0,114,178} 
\definecolor{ego_color}{HTML}{F5615C} 
\definecolor{dsafe_color}{HTML}{8770FE} 

\definecolor{concrete_color}{HTML}{8bd5ff}

\definecolor{symbolic_color}{HTML}{ec39b7} 

\definecolor{mc_color}{HTML}{5a2ecb} 

\definecolor{goal_color}{RGB}{0,158,115} 

\definecolor{nnv_color}{RGB}{0,158,115} 
\usepgfplotslibrary{groupplots}

\usetikzlibrary{arrows}
\tikzset{>=stealth'}
\tikzset{every picture/.style={font issue=\footnotesize},
         font issue/.style={execute at begin picture={#1\selectfont}}
        }

\definecolor{mycolor1}{rgb}{1.00000,0.60000,0.00000}%
\definecolor{mycolor2}{rgb}{0.80000,1.00000,0.00000}
\definecolor{pastelGreen}{HTML}{3EAA0D}

\definecolor{pastelMagenta}{HTML}{FF48CF} 
\definecolor{pastelPurple}{HTML}{8770FE} 
\definecolor{pastelBlue}{RGB}{0,114,178} 
\definecolor{pastelSkyBlue}{RGB}{86,180,233} 
\definecolor{pastelSeaGreen}{RGB}{86,180,233} 
\definecolor{pastelGreen}{RGB}{0,158,115} 
\definecolor{pastelOrange}{RGB}{230,159,0} 
\definecolor{pastelRed}{HTML}{F5615C} 
\definecolor{darkColor}{HTML}{300A24} 

\usepackage{multirow}
\usepackage{booktabs}
\usepackage{lscape}

\usepackage{optidef}

\newcommand{\x}{\mathbf{x}}
\newcommand{\vect}[1]{\mathbf{#1}}

\newcommand{\fakeparagraph}[1]{\indent {\emph{#1}. }}

\ShortHeadings{Safety Verification of Neural Network Control Policies for Nonlinear Systems}
{Sidrane, Maleki, Irfan \& Kochenderfer}
\firstpageno{1}

\begin{document}

\title{{OVERT}: An Algorithm for Safety Verification \\ of Neural Network Control Policies for Nonlinear Systems}

\author{\name Chelsea Sidrane*\email csidrane@stanford.edu \\
       \name Amir Maleki* \email amir.maleki@stanford.edu \\
       \name Ahmed Irfan \email irfan@cs.stanford.edu \\
       \name Mykel J. Kochenderfer \email mykel@stanford.edu \\
       \addr Aeronautics \& Astronautics Department, Stanford University \\
        496 Lomita Mall, Stanford, CA 94305 USA
        \\
        (* denotes equal contribution)
        }
\editor{}


\maketitle

\begin{abstract}%
Deep learning methods can be used to produce control policies, but certifying their safety is challenging.
The resulting networks are nonlinear and often very large.
In response to this challenge, we present OVERT: a \emph{sound} algorithm for safety verification of nonlinear discrete-time closed loop dynamical systems with neural network control policies.
The novelty of OVERT lies in combining ideas from the classical formal methods literature with ideas from the newer neural network verification literature. 
The central concept of OVERT is to abstract nonlinear functions with a set of optimally tight piecewise linear bounds.
Such piecewise linear bounds are designed for seamless integration into ReLU neural network verification tools. 
OVERT can be used to prove bounded-time safety properties by either computing reachable sets or solving feasibility queries directly. 
We demonstrate various examples of safety verification for several classical benchmark examples. 
OVERT compares favorably to existing methods both in computation time and in tightness of the reachable set.
\end{abstract}

\section{Introduction} \label{sec:intro}
Deep learning has found application in a wide variety of fields, from computer vision and natural language processing to control of autonomous agents. 
In particular, recent innovations in deep reinforcement learning have yielded impressive results, such as training neural networks to play Atari video games \citep{mnih2015human} and board games like Go \citep{silver2016mastering}. 
As a result, there is increasing interest in applying deep learning to safety-critical control problems such as autonomous driving~\citep{chen2015deepdriving,
bojarski2016end} and aircraft collision avoidance~\citep{julian2016policy}. 
Although neural networks can provide flexible representations of complex control strategies, there is concern that they can result in unexpected behavior~\citep{szegedy2013intriguing, papernot2016limitations}. 

Testing can play an important role in identifying undesirable behavior in a system, but testing alone is not sufficient to prove the absence of failures.
It is possible that rare, catastrophic failures exist that intelligent testing schemes are required to discover~\citep{corso2020survey}. 
Instead, formal verification can be used on a model of the system in order to prove the absence of failures.
More specifically, in this work, we develop an algorithm to prove or disprove safety properties of nonlinear discrete-time dynamical systems with deep neural network control policies. 
The incorporation of nonlinear functions as well as neural networks makes these systems challenging to verify. 
\citet{katz2017reluplex} demonstrated that verifying ReLU neural networks alone is an NP-complete problem.
Tools like those developed by \citet{cimatti2018incremental} can verify properties of discrete-time systems with nonlinear functions.
However, these tools cannot efficiently handle the complexity of neural networks, which can contain thousands of smooth nonlinear or piecewise-linear activation functions~\citep{katz2017reluplex}.
This problem has led to the development a family of neural network verification tools that are able to efficiently verify properties of neural networks~\citep{liu2020algorithms}.
Together with a small but growing body of research, this work aims to bridge the gap between the existing literature on verification of closed-loop systems and literature on neural network verification. 

There is extensive literature for formal verification of discrete-time closed-loop systems (also known as \textit{transition systems})~\citep{DBLP:reference/mc/2018}.
Much of this work focuses on linear systems. 
Although nonlinear model checkers such as the tool developed by \citet{cimatti2018incremental} can handle nonlinear dynamics, they are not designed for use with neural networks. 
Even if a network can be encoded into the tool, which may not be possible, the large number of nonlinear activation functions present difficulties.
For example, if the network has ReLU activation functions, each would be expanded eagerly to a disjunction, leading to an intractable runtime.

The inability of existing model checking tools or other automated reasoning tools to verify neural networks has led to a recent surge in development of verification tools that are capable of reasoning about neural networks with ReLU activation functions~\citep{katz2017reluplex, liu2020algorithms}. 
When the network that we would like to verify represents a control policy, these tools may be used to verify input-output properties of the control policy in isolation. 
However, verifying properties of the control policy in isolation does not address the entire closed-loop system of which the control policy is just one part. 
Our work uses these recent developments in neural network verification in order to reason about the closed-loop system. 

There is prior work that also uses new neural network verification tools to reason about the closed-loop system. 
Some of these works can be used for discrete-time systems~\citep{tran2020nnv3, julian2021reachability}. 
The methods presented in both \citet{tran2020nnv3} and \citet{julian2021reachability} work by computing polytopes containing the reachable set of the control policy and the dynamics at each timestep in an iterative fashion. 
These methods are complementary to ours, but have two main limitations. 
The first is that they are based on hybrid systems reachability tools, which scale poorly with the number of state variables \citep{Flowstar}. 
The second is that the underlying reachability algorithm based on iterative computation of concrete reachable sets leads to a compounding looseness of the approximation, known as the wrapping effect \citep{neumaier1993wrapping}. 
To fix this, the initial set may be split into smaller cells across each dimension, but this too can lead to poor scaling in the number of state variables. The methods developed here, in contrast, have either no explicit dependence on the number of state variables or only linear dependence. All methods presented mitigate the wrapping effect by preserving a symbolic representation across multiple timesteps. 

There is related work that assumes piecewise-linear system dynamics, or approximates (neither underapproximates nor overapproximates) potentially nonlinear system dynamics using data-driven models~\citep{dutta2018learning, akintunde2018reachability, akintunde2020formal, xiang2018piecewise}. However, if the system dynamics are approximated with a data driven model, proving that there are no counterexamples for the approximate system does not allow a sound claim that there are no counterexamples for the original nonlinear dynamics. Our method makes an overapproximation that \emph{does} allow such a sound claim to be made. Finally, there is adjacent work that addresses continuous-time systems~\citep{ivanov2018verisig, huang2019reachnn, dutta2019reachability}. These methods are closely related but not directly applicable to the problem at hand.

\fakeparagraph{Contributions}
We present a method that we call OVERT that reasons about nonlinear discrete-time closed-loop systems that contain neural network control policies.
OVERT overapproximates nonlinear dynamical systems using piecewise-linear relations, making the entire closed-loop system a conjunction of piecewise-linear relations. 
The resultant system of piecewise-linear relations can be efficiently reasoned about using ReLU neural network verification tools.
The experiments we present are encoded and solved as a mixed-integer programs.
We prove bounded-time safety properties for the closed-loop system by unrolling the system in time. 
Reachability queries can be solved directly as feasibility problems or by explicitly computing the reachable set.
The use of hybrid-symbolic reachable set computation preserves both tightness of the reachable set and tractability of its computation.
The methods presented have at most linear explicit dependence on the number of state variables.
The use of an overapproximation allows the sound claim that a proof of no counterexamples for the approximation implies no counterexamples for the original system. 
We have released two Julia packages, OVERT.jl and OVERTVerify.jl, which provide implementations of the sound overapproximation and verification algorithms, respectively.


\section{Background}
This paper borrows ideas from formal verification, control theory, and machine learning. This background section reviews prerequisite concepts and terminology. 

\subsection{Closed-Loop Systems}
A \emph{closed-loop system} is a \emph{dynamical system} paired with a \emph{feedback control policy}. 
The term \emph{dynamical system} originates in control theory and describes a model of a stateful system that evolves in time. 
To illustrate, we use a system of an inverted pendulum as a running example. 
The state, $\vect x$, of the inverted pendulum may be represented by the angle of the pendulum, $\theta$, and the angular velocity of the pendulum, $\dot \theta$: $\vect x = [ \theta , \dot \theta ]^T$.
The set of all possible states is $[-\pi, \pi] \times \mathbb R$. We can also  define an initial set of states, also known as an initial condition.
A dynamical system evolves in time according to an update function, which is commonly called the equation of motion or the plant in control theory.
A dynamical system may be discrete-time or continuous-time. 
We focus on discrete-time systems. 
For discrete-time systems, the update function $\vect h$ takes as input the state at time $t$ and returns the state at time $t+1$: $\vect x _{t+1} = \vect h(\vect x_t)$. 
Additionally, a dynamical system may be linear or nonlinear. A nonlinear system has a nonlinear update function $h$. 
In this work, we consider nonlinear systems where $h$ may contain polynomials, transcendentals ($\sin$, $
\cos$, exp), piecewise-linear functions, and compositions thereof. The dynamical systems that arise in classical mechanics can typically be expressed using a function in this class.

A  dynamical system may also incorporate external inputs, called control inputs, that allow us to generate desired behavior. 
The update function is then $\vect x _{t+1} = \vect f(\vect x_t, \vect u_t)$ where $\vect u_t$ is the control input at time $t$ produced by the \emph{control policy}. 
The control policy is also called the controller in control theory or the policy in reinforcement learning.
In the inverted pendulum example, a control input could be torque applied at the pivot joint. This torque may be used to produce desired behavior, such as balancing the pendulum up-right. 
A \emph{feedback} control policy produces control input $\vect u_t$ as a function of the current state, $\vect u_t = \vect c(\vect x_t)$, making it reactive to changes in the system.
The prevalence of deep learning and deep reinforcement learning has led to the use of neural network based control policies.
Consequently, this work focuses on nonlinear discrete-time closed-loop systems with neural network control policies. 

\subsection{Transition Systems}
Readers from formal verification and model checking may be familiar with the notion of a \emph{transition system}. 
A discrete-time dynamical system, paired with an initial condition (initial set of states), may  be equivalently modeled as a symbolic transition system. 
A symbolic transition system has states that evolve in discrete time according to a transition relation, beginning from a specific initial set. 
Such a symbolic transition system is defined by a tuple $(X, I, TR)$, where $X$ is a set of state variables.
Here, $I(X)$ represents the set of initial states using a formula over the variables in $X$. 
A \emph{formula} is a Boolean combination (using standard logical operators) of constraints. 
A constraint is of the form $p \bowtie 0$, where $p$ is a polynomial (summation of monomials) and $\bowtie \in \{\leq, <, =, \geq, >\}$.
A \emph{transition relation} $TR$ describes how the state evolves in time. 
The transition relation is symbolically defined as a formula over the state variables associated with the current and next time step.
Let $\vect x_t$ be the current state at time $t$ and $\vect x_{t+1}$ be a next state at time $t+1$.
The formula
$TR(\vect x_t, \vect x_{t+1})$ returns true if state $\vect x_{t+1}$ is a valid successor of state $\vect x_t$.
A transition system does not require that there be a single valid successor state for a given state.
The update function of a dynamical system corresponds to the transition relation when modeling a dynamical system as a transition system.

\subsection{Proving Properties}
We prove safety properties and goal-reaching properties for closed-loop systems, which can equivalently be described as model checking for transition systems; specifically, bounded-time model checking. 
For readers from control theory, this can also be described as solving discrete-time \textit{reachability problems}. 
A reachability problem reasons about whether the set of states reachable over time intersects with an unsafe set, or reaches a goal set.
For example, consider modeling the motion of an inverted pendulum as a transition system or discrete-time dynamical system.
The properties we prove can be conditions that we would like to always hold, known as invariant properties,
e.g. \textit{it is always true that the angle of the pendulum with respect to vertical is greater than \SI{-5}{\degree}}, or conditions that we would \textit{eventually} like to hold, such as \textit{the angle of the pendulum with respect to vertical is eventually greater than \SI{-5}{\degree}}. 
Formally, a property is defined as a Boolean-valued predicate over the states of the system, e.g. $\theta \geq \SI{-5}{\degree}$, which is then combined with a \textit{temporal modal operator} such as $G$ (Globally) or $F$ (Finally) as well as a finite range of timesteps to form a bounded temporal property. 
For example, $F_{1:10}(\theta \geq \SI{-5}{\degree})$ expresses that $\theta$ must be greater than $\SI{-5}{\degree}$ at some point in the first 10 timesteps. 
The notation $G_{1:10}(\theta \geq \SI{-5}{\degree})$ expresses that $\theta$ must be greater than $\SI{-5}{\degree}$ at all steps in the first 10 timesteps. 
OVERT has the capability to reason about a larger fragment of temporal logic than just the $F$ and $G$ operators. In this paper, the $G$ operator is used to express safety properties, and the $F$ operator is used to express goal-reaching properties.

In order to prove that a property (e.g., $G_{1:10} \phi$) holds, the negation of the property must be proven unsatisfiable (UNSAT). 
If a \emph{trace} of the system is found that satisfies the negation of the property, this trace is a counter-example demonstrating where the property is violated.
A trace may also be described as a trajectory in the control theory context. 
The $F$ and $G$ operators are dual, meaning that $\neg G \phi$ is equivalent to $F \neg \phi$ as well as $\neg F \phi$ is equivalent to $G \neg \phi$. 
In order to prove $G_{1:10} \phi$ holds, we would test whether the complement, $F_{1:10} \neg \phi$, is unsatisfiable.
At each timestep $t \in 1:10$, the property $\neg \phi$ should be unsatisfiable.

\section{Methods} \label{sec:method}

Consider a closed-loop discrete-time dynamical system with update function:
\begin{equation*}
    \mathbf{x}_{t+1} = \vect{h}(\mathbf{x}_t)
\end{equation*}
where $\mathbf{x}_t = \left[x_{t,1}, x_{t,2}, \cdots, x_{t,n}\right] \in \mathcal{X} \subseteq \mathbb R^n$ denotes the state vector at time $t$ and $\vect{h}: \mathbb R^n \rightarrow \mathbb R^n$ denotes the state update function. 
The closed-loop update function $\vect{h}$ is comprised of the dynamics function $\vect{f}(\x, \vect{u})$ and the control policy function $\vect{u} = \vect c(\x)$:
\begin{equation*}
    \vect{h}(\x) = \vect{f}(\x, \vect{c}(\x))
\end{equation*}
For now, we assume that the control policy $\vect{c}(\x)$ is represented by a neural network with piecewise linear activation functions (e.g., ReLU). 
We will discuss how other activation functions may be used in \cref{sec:exten}. 
The dynamics function $\vect{f}$ may contain nonlinearities that prohibit the application of neural verifications tools such as ReluPlex \citep{katz2017reluplex}, MIPVerify \citep{tjeng2017evaluating}, and ReluVal \citep{wang2018formal} because these tools only support linear and piecewise linear operations.
To circumvent this problem, 
we construct an \emph{abstraction} of the dynamics function that only contains linear and piecewise linear relations (see \cref{sec:abstrac}).
The abstraction of the dynamics function using linear and piecewise linear relations corresponds to creating an overapproximation of the set of values forming the image of the update function that varies based upon the input set.

Once the dynamics have been abstracted, the closed-loop system is unrolled in time.
Reachability queries can then be posed to assess safety properties or goal-reaching properties (see \cref{sec:problemtypes}).
The resulting system consists only of piecewise-linear constraints and is suitable for translation into many ReLU neural network verification tools. 
In this work, the unrolled system is encoded as a mixed integer program (see \cref{sec:solve-mip}), which is the basis of several ReLU neural network verification tools~\citep{tjeng2017evaluating, akintunde2018reachability}.
The mixed integer program is then solved with Gurobi (\cite{gurobi}). 

\subsection{Constructing Multidimensional Overapproximations}
\label{sec:abstrac}
Many modern nonlinear model checking methods involve \emph{abstraction} of nonlinear relations and functions.
Abstraction is the process of deriving an approximate representation of the transition system that is more computationally tractable for verification. 
Importantly, an abstraction is constructed such that proving certain properties of the \emph{abstracted} system automatically proves that the same properties hold for the original system, also called the \emph{concrete system}.
Such an abstraction is also called an \emph{overapproximation} 
because the set of problem states that the abstracted system may visit, $\mathcal R_A$, is a superset of the set that the original system may visit, $\mathcal R_O \subseteq \mathcal R_A$.
In this work, we abstract the closed-loop system using a \emph{relational overapproximation} (also known as a \emph{relational abstraction}). The details of how this is done will be explained in this section.  

OVERT can overapproximate functions with both multidimensional inputs and multidimensional outputs. 
The update function $\vect f$ may be represented as a vector of functions $[f_1,f_2,\ldots,f_n]^T$ each of which maps $\mathbb R^n \rightarrow \mathbb R$.
The update function for the $i$th component of the state $\vect x_{t,i}$ is denoted $\vect x_{t+1,i} = f_i (\vect x_t, \vect c (\vect x_t))$. 
The algorithm presented is applied to each update function component $f_i$ separately. The subscript is dropped for simplicity in the description below. 
The algorithm consists of two main steps: rewriting $\vect f$ from a single complex equation into a series of simpler constraints and then overapproximating any nonlinear constraints.
Both the rewriting and approximation steps are described below.

\subsubsection{Rewriting}
We rewrite the dynamics function  as a conjunction of constraints (relations), where each constraint is either an elementary function $e(x)$ (such as $\sin(x)$, $\log(x)$, $\exp(x)$, $x^3$), an algebraic binary operation (addition ($+$) or subtraction ($-$)), or a single-juncture piecewise linear function (such as $\max(x, y)$, $\min(x, y)$).
Many common dynamics functions that arise in classical mechanics can be rewritten this way by introducing auxiliary variables. 
For example, consider the function 
\begin{equation*}\label{eq:f_example}
    f(x,y,z) = \sin(x^2 + y - \log z) ~~~\text{defined over}~~~ \mathcal{X}:= \{(x,y,z) \mid 1\leq x,y, z \leq 2\}
\end{equation*}

The first column of \cref{table:f_example} shows the outcome of the rewriting step, where function $f$ is split into a number of linear and nonlinear equality and non-equality relations. \Cref{alg:overtstepone} 
specifies \emph{how} this step is executed iteratively.
On each iteration, we isolate the operator and operands using the function \Call{Split}{}. 
Depending on the type of operator, the algorithm proceeds. 
Apart from scalar multiplication, we cannot use multiplication ($\times$) and division ($\div$) operations because it will lead to nonlinear constraints in the approximation step. Therefore, the function \Call{ConvertMultiplication}{} converts these operations into an exponential of sum of logarithms using the following two identities:
\begin{align*}
x \times y &= \exp \left(\log(x) + \log(y) \right), ~~~~ \forall x, y >0 \\
 x / y &= \exp \left(\log(x) - \log(y) \right), ~~~~ \forall x, y >0  
\end{align*}

A transformation of variables on $x$ and $y$ is used so that the domain for each $\log$ is $[\xi, 1+\xi]$, where $\xi$ is a small positive constant, so that $\log$ may be applied. 

\begin{algorithm}
\begin{algorithmic}[1]
\Function{Rewrite}{expr, container}
\If{expr is a variable or number}
    \State \Return expr, container
\ElsIf{expr is affine} 
    \State $v \gets$ generate new variable
    \State Add ($v = $ expr) to container 
    \State \Return $v$, container
\ElsIf{operator of expr is $\times$ or $\div$}
    \State Convert expr using $\log$ and $\exp$
    \State \Return \Call{Rewrite}{expr, container}
\ElsIf{operator of expr is $+$ or $-$}
    \State operator, operands = split(expr)
    \State $x$, container $\gets$ \Call{Rewrite}{operands[1], container}
    \State $y$, container $\gets$ \Call{Rewrite}{operands[2], container}
    \State $v \gets$ generate new variable
    \State Add (operator($x$,$y$) = $v$) to container
    \State \Return $v$, container
\ElsIf{ operator is a supported nonlinear unary function}
    \State operator, operand $\gets$ split(expr)
    \State $x$, container $\gets$ \Call{Rewrite}{operand, container}
    \State $v \gets$ generate new variable
    \State Add (operator(x) = $v$) to container
    \State \Return $v$, container
\EndIf
\EndFunction
\end{algorithmic}
\caption{Rewriting \label{alg:overtstepone}}
\end{algorithm}

\subsubsection{Approximation}
This step is comprised of processing the constraints obtained from the re-writing step that contain nonlinear functions.
For each nonlinear equation such as $v_1 = \sin(x)$, the equation is replaced with two relations forming upper and lower bounds: $v_1 \leq \sin_{UB}(x)$ and $v_1 \geq \sin_{LB}(x)$. Because each nonlinear \textit{equation} is replaced with two piecewise linear \textit{relations} we refer to the result as a \emph{relational overapproximation}. The second column in \cref{table:f_example} shows the result of the approximation step on an example.

\Cref{alg:overtsteptwo} specifies how this process is performed. The two functions \Call{GetUpperBound}{} and \Call{GetLowerBound}{} compute piecewise linear upperbound and lowerbound functions, respectively.
More precisely, for each nonlinear elementary function $e(x)$ defined over the interval $[a,b]$, functions \Call{GetUpperBound}{} and \Call{GetLowerBound}{} compute $e_{UB}$ and $e_{LB}$, respectively, such that
\begin{equation*}\label{eq:bound_definition}
    e_{LB}(x) \leq e(x) \leq e_{UB}(x) ~~~ \forall x \in [a,b]
\end{equation*}

\begin{algorithm}
\begin{algorithmic}[1]
\Function{Approximate}{container, parameters}
\State approximation $\gets \emptyset$
\For{($v$ = expr) in container}
    \If{expr is nonlinear}
        \State Add ($v \leq$ \Call{GetUpperBound}{expr, parameters}) to approximation
        \State Add ($v \geq$ \Call{GetLowerBound}{expr, parameters}) to approximation
    \Else 
        \State Add ($v$ = expr) to approximation
    \EndIf
\EndFor
\State \Return approximation
\EndFunction
\end{algorithmic}
 \caption{Approximation \label{alg:overtsteptwo}}
\end{algorithm}

\begin{table}[h]
\caption{OVERT applied to $f$ defined in \cref{eq:f_example}.  \label{table:f_example}}
\begin{center}
\begin{tabular}{@{}l|l@{}}
\toprule
\textbf{Step 1}          & \textbf{Step 2} \\ \midrule
$ f(x,y,z) = v_{11} $       & $f(x,y,z) \approx v_{11}$              \\
$v_{11} = \sin(v_8) $       & $v_{11} \leq v_9$  \\
                            & $v_{11} \geq v_{10} $ \\
                            & $v_9 = e_{UB_{v_{11}}}(v_8)$               \\
                            & $v_{10} = e_{LB_{v_{11}}}(v_8)$      \\
$v_8 = v_4 -v_7$            & $v_8 = v_4 -v_7$     \\
$v_7 = \log(z)$              & $v_7 \leq v_5$    \\
                            & $v_7 \geq v_6 $   \\
                          & $v_5 = e_{UB_{v_{7}}}(z)$      \\
                          & $v_6 = e_{LB_{v_{7}}}(z)$      \\
 $v_4 = v_3 + y$          &  $v_4 = v_3 + y$      \\
$v_3 = x^2 $             & $v_3 \leq v_1$                 \\
                         & $v_3 \geq v_2$                 \\
                         & $v_1 = e_{UB_{v_{3}}}(x)$     \\
                         &  $v_2 = e_{LB_{v_{3}}}(x)$\\
\bottomrule
\end{tabular}
\end{center}
\end{table}
 
The algorithm for finding the upper-bound piecewise linear function ($e_{UB}$) and the lower-bound piecewise linear function ($e_{LB}$) used in \Call{GetUpperBound}{} and \Call{GetLowerBound}{} is described in \cref{sec:oneD} and \cref{sec:closed-form}, with a derivation in \cref{sec:app1}. 
This step is important because the overapproximation is useful only if it is \emph{practically} tight; i.e. the approximation error is small. 
Otherwise, OVERT may be unnecessarily conservative and generate spurious counter examples. 
A spurious counter example is when the solver reports SAT and returns a point in state space where the property does not hold for the abstracted system, but the property does hold for the true system. 
However, reducing approximation error requires more linear segments in the piecewise linear bound, which leads to a more computationally expensive representation. 
In \cref{sec:app1}, we derive an algorithm to find optimally tight bounds $ e_{LB}(x)$ and $ e_{UB}(x)$ for any one-dimensional function $e(x)$.
This tight bound is efficient in its use of piecewise linear constraints.

The relational overapproximation of the dynamics together with the ReLU neural network control policy forms a system of linear and piecewise linear constraints.
Such a system can be analyzed using recent algorithms introduced for ReLU neural network verification. 
As we will discuss in \cref{sec:problemtypes}, we solve this system using an approach inspired by the MIPVerify algorithm \citep{tjeng2017evaluating}. We initially prototyped using the Reluplex algorithm \citep{katz2017reluplex}, but the MIP formulation proved faster.

\subsection{Constructing One-Dimensional Overapproximations} \label{sec:oneD}
The upper and lower bounds for each one-dimensional nonlinear function $f$ 
described in the previous section are constructed by first separating a given function into convex and concave regions.
To ensure continuity across regions of differing convexity, we enforce that the endpoints of a bound for a region of uniform convexity must be coincident with the function $f$.
The current implementation can compute the convex and concave intervals of the following set of unary functions: $\{\cos, \sin, \exp, \log, \tanh, c/x, c^x, x^c\}$. 
This set of functions is expressive, however support for additional functions may be easily added. 
Two additional automated options also exist: implementing symbolic differentiation and using a certified interval root finding method, and independently verifying the overapproximations generated by OVERT. 

Once the convex and concave regions of a function have been obtained, the process of constructing the lower bound for a convex region and the upper bound for a concave region are homologous, as is the process of constructing an upper bound for a convex region and a lower bound for a concave region. 
Both are described below.

\subsubsection{Upper Bound for a Convex Region}
\label{sec:bnd_convex}
\begin{figure}
\centering
\includegraphics[width=0.6\textwidth]{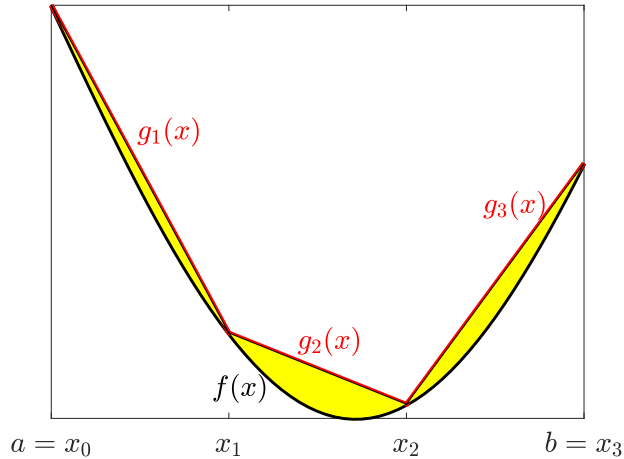}
\caption{A convex function $f(x)$ and its upper bound $g(x)$, which is composed of $n=3$ linear pieces: $g_1(x), g_2(x)$, and $g_3(x)$.}
\label{fig:fig_g_convex}
\end{figure}

An upper bound $g(x)$ for a convex region of the function $f(x)$ can be formed using the secant line connecting the ends of the interval, $a$ and $b$, which follows from the definition of convexity. However, each convex interval is divided into $n$ subintervals, $[x_{i-1}, x_i]$, for $i \in 1:n$, and the secant line $g_i(x)$ connecting the endpoints of the subinterval is used to bound the function. To ensure a tight bound, the placement of intermediate points $x_i$ are optimized to obtain the upper bound that minimizes the area between the bound and the function:
\begin{equation*}
    \min_{x_1:x_{n-1}} \int_a^b \left[g(x; \mathbin{x_{0:n}}) - f(x) \right] \mathrm{d}x
\end{equation*}
where the endpoints, $g(x_0=a)=f(a)$ and $g(x_n=b)=f(b)$, are fixed.
Evaluating this expression yields the optimality condition:
\begin{align} \label{eq:convex_optim_body}
f'(x_i) = \frac{f(x_{i+1}) -f(x_{i-1})}{x_{i+1} -x_{i-1}}
\end{align}
In other words, the slope of $f$ at point $x_i$ must match the slope of a secant connecting points $x_{i-1}$ and $x_{i+1}$. A derivation can be found in \cref{sec:app-conv}.
Points $(x_i)$ that satisfy the optimality conditions in \cref{eq:convex_optim_body} are found using NLsolve.\footnote{\url{https://github.com/JuliaNLSolvers/NLsolve.jl}}
The values $y_i$ are specified by the function:
\begin{equation*}
    y_i = f(x_i)
\end{equation*}
If the numerical routine incurs errors in selecting points $x_i$ this may affect the optimality of the bound but will not affect the validity, as any secant connecting points $(x_{i-1}, f(x_{i-1}))$ and $(x_i,f(x_i))$ is a valid upper bound over the interval $[x_{i-1}, x_i]$.
An illustration of a possible bound $g(x)$ is shown in \cref{fig:fig_g_convex}.

\subsubsection{Upper Bound for a Concave Region}
\label{sec:bnd_concave}
An upper bound $g(x)$ for a concave region of the function $f(x)$ can be formed using any tangent line, which comes from the definition of convexity.
However, using a single tangent line to bound the entire extent of a concave region of the function would not result in a tight bound.
Consequently, each concave interval is divided into $n$ subintervals, $[x_{i-1}, x_i]~,~i\in 1:n$, and a tangent line $g_i(x; \alpha)$ is used to bound the function over each subinterval, where $\alpha$ is the point of tangency.
To ensure a tight bound, the parameters of this piecewise linear bound are optimized. For any given subinterval $[x_{i-1}, x_i]$ bounded by a single tangent line segment $g_i(x;\alpha)$, placing the tangent point $\alpha$ at the midpoint of the subinterval yields the tightest bound. \Cref{sec:app0} provides a proof.

OVERT also optimizes the placement of the points $x_i~,~i\in 1:n-1$ that divide the concave interval into subintervals. The points $x_i$ are optimized to obtain an upper bound that minimizes the area between the bound and the function:
\begin{equation*}
\min_{x_1:x_{n-1}} \int_a^b \left[g(x; \mathbin{x_{0:n}}) - f(x) \right] \mathrm{d}x
\end{equation*}
Minimizing this objective yields an optimality condition for the $x_i$:
\begin{align}
    x_i &= h \left(\frac{x_{i-1}+x_i}{2}, \frac{x_i + x_{i+1}}{2} \right) \label{eq:optim1} \\
    h(\alpha, \beta) &= \frac{\beta f'(\beta)-\alpha f'(\alpha)}{f'(\beta)-f'(\alpha)} - \frac{f(\beta)-f(\alpha)}{f'(\beta)-f'(\alpha)} \label{eq:h}
\end{align}
This condition implies that the optimal bound is continuous within a region of uniform concavity: $g_i(x_i) = g_{i+1}(x_i)$. 
Proof as well as deriviation of equations \cref{eq:optim1} and \cref{eq:h} is provided in \cref{app:overapprox1D}.
In order to ensure continuity across regions of different concavity, the bound is constrained to be tangent at the endpoints: $[a,b]$. Consequently, the bound is given by points $x_i$ that satisfy:
\begin{subequations}
\label{eq:optim}
\begin{align}
x_0 =&~ a \\
x_1 =&~ h\left(a, \frac{x_1 + x_{2}}{2} \right)\\
x_i =&~ h \left(\frac{x_{i-1}+x_i}{2}, \frac{x_i + x_{i+1}}{2} \right),~~~ i\in \{2, \cdots, n-2\}\\
x_{n-1} =&~ h \left(\frac{x_{n-1}+x_n}{2}, b \right) \\
x_{n} =&~ b
\end{align}
\end{subequations}
where $h$ is given by \cref{eq:h}.
The $y_i$ values are given by the line that is tangent at the midpoint of the $i$th interval,
\begin{equation*}
g_i(x) = f'\left(\frac{x_{i-1}+x_i}{2}\right) \left(x - \frac{x_{i-1}+x_i}{2}\right) + f\left(\frac{x_{i-1}+x_i}{2} \right)
\end{equation*}
except for the first two and last two $y_i$:
\begin{subequations}
\begin{align*}
    y_0 =& f(a) \\
    y_1 =& f'(a)(x_1 - a) + f(a) \\
    y_i =& f'\left(\frac{x_{i-1} + x_i}{2}\right)\left(\frac{x_i - x_{i-1}}{2}\right) + f\left(\frac{x_{i-1} + x_i}{2}\right), ~i=2,...,(n-2) \\
    y_{n-1} =& f'(b)(x_{n-1} - b) + f(b) \\
    y_n =& f(b)
\end{align*}
\end{subequations}

In practice, we use a numerical nonlinear solver to produce points $x_i$ satisfying the optimality constraints in~\cref{eq:optim}. 
We then check the solution to ensure that the optimality conditions have been met. 
If the optimality conditions have not been met, the resultant bound is still sound, but may not be continuous over the interval. In order to restore continuity, the bound may be repaired by taking $y_i$ to be $\max(g_i(x_i), g_{i+1}(x_i))$. 
Any line segment adjusted in this way is greater than or equal to the original bound $g_i(x)$ for $x\in [x_{i-1},x_i]$ and therefore still a valid upper bound.  For a lower bound for a convex function, the bound may be analogously repaired by taking $y_i = \min(g_i(x_i), g_{i+1}(x_i))$.

\subsubsection{Implementation Details}
Once the bounds have been constructed, an $\epsilon$ gap is added to each bound such that the bounds do not ``touch'' the original function. All points $(x_i, y_i)$ that make up the the upper bound are shifted by $+\epsilon$ and all points $(x_i, y_i)$ that make up the lower bound are shifted by $-\epsilon$. This gap helps ensure that any errors incurred during floating point computations do not compromise the validity of the bounds. 

\subsubsection{Examples}
\begin{figure}[t]
\centering
\input{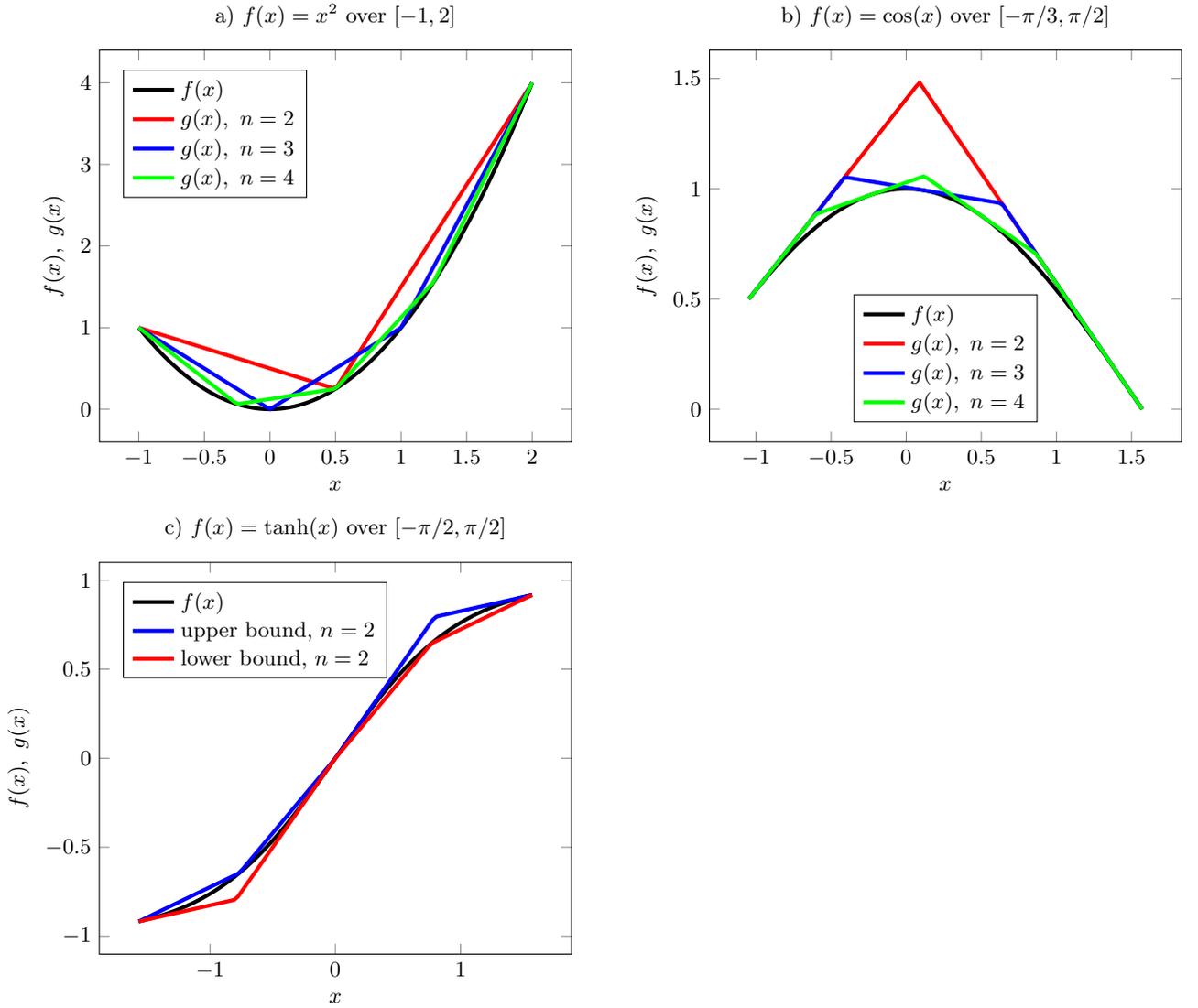}
\caption{\label{fig:fig_1d} Examples of piecewise linear over-approximations. }
\end{figure}

\Cref{fig:fig_1d} shows three different functions with their associated upperbound function $g(x)$. 
In \cref{fig:fig_1d}a, $f(x) = x^2$ is plotted in black over the interval $[a,b]=[-1,2]$. The function $f(x)=x^2 $ is convex. 
Three upper bound function $g(x)$ are shown in color for three different values of $n$. 
One may notice that the accuracy of approximations improves significantly when $n$ increases. 
In \cref{fig:fig_1d}b, we repeat this example for $f(x) = \cos(x)$ which is concave over the interval $[a,b]=[-\pi/3, \pi/2]$. 
Finally, in \cref{fig:fig_1d}c, the function $f(x) = \tanh(x)$ is bounded over the interval $[a,b]=[-\pi/2,\pi/2]$. 
In this case, $f(x)$ has an inflection point at $x=0$, which means the interval is split into two sub-intervals $[-\pi/2,0]$ and $[0, \pi/2]$. 
The blue and red curves show the upper bound and the lower bound of the function, respectively. 
In both cases, $n=2$ linear segments were used over each sub-interval.
Notice that both the upper and lower bound are continuous over $[-\pi/2,\pi/2]$. 

\subsection{Closed Form Expressions of Piecewise Linear Functions} \label{sec:closed-form}


The final step is to transform the set of points $(x_i, y_i)$ representing the lower and upper bound into closed form expressions. 
While each individual piece of a bound $g_i(x)$ could be written as a line segment connecting points $(x_i, y_i)$:
\begin{equation*}\label{eq:g(x)_body}
g_i(x) = \frac{y_i - y_{i-1}}{x_i - x_{i-1}} \left( x-x_i\right) + y_i, ~~~ x_{i-1} \leq x \leq x_i
\end{equation*}
representing the bounds in closed form is useful because we can then write the overapproximation of a nonlinear function using two linear inequalities: $e_{\sin_{LB}}(x) \leq \sin(x) \leq e_{\sin_{UB}}(x)$.
The closed-form expression must satisfy the following two conditions: i) $g(x)$ is piecewise linear and continuous and ii) $g(x_i) = y_i$ for $n+1$ points $(x_i, y_i)$. 
The points are already computed following either algorithm in \cref{sec:oneD}. 
The function $g(x)$ is defined to be a weighted sum of $n+1$ piecewise linear basis functions $\beta_i(x)$ with the \emph{property} that:
\begin{equation*}
\beta_i(x_j) = \begin{cases} 
      1 & \text{if } i = j \\
      0 & \text{if } i \neq j 
   \end{cases}    
\end{equation*}
Then, we can write the function $g(x)$ as: 
\begin{equation*} \label{eq:closed_form_g}
	g(x) = \sum_{i=0}^n \beta_i(x) y_i
\end{equation*}
Notice that $g(x)$ is piecewise linear because it is a weighted sum of piecwise linear functions $\beta_i(x)$. In addition, 
\[g(x_i) = y_0\beta_0(x_i) + \cdots + y_i\beta_i(x_i) + \cdots + y_N\beta_N(x_i) = 0 + \cdots + y_i + \cdots + 0 = y_i \]

The problem is therefore reduced to finding the basis functions $\beta_i(x)$. Our idea for constructing the basis functions is inspired by first order basis functions in finite element analysis. The basis functions $\beta_i(x)$ can be defined using the following piecewise relation:
\begin{equation}\label{eq:beta_i_pw}
    \beta_i(x) = 
    \begin{cases} 
      0 & x < x_{i-1} \\
      L^i_l & x_{i-1} < x < x_i \\
      L^i_r & x_i < x < x_{i+1} \\
      0 & x > x_{i+1} 
   \end{cases}
\end{equation}
where
\begin{equation*}
L^i_l = \frac{x-x_{i-1}}{x_i - x_{i-1}}, ~~~~ L^i_r = \frac{x-x_{i+1}}{x_i - x_{i+1}}
\end{equation*}
\Cref{eq:beta_i_pw} can be rewritten in closed form as:
\begin{equation*} \label{eq:alpha_i}
\beta_i(x) = \text{min}\Big(\text{max}\big(L^i_{l},0\big),  \text{max}\big(L^i_{r},0\big) \Big) = \text{max}\Big(0, \text{min} \left(L^i_l, L^i_r \right) \Big)
\end{equation*} 

For the boundary cases of $i=0$ and $i=n$, we only have $L_r^0$ and $L_l^n$, respectively, and therefore, 
\begin{equation*} \label{eq:hi_boundary}
\beta_0 = \text{max}\big(L^0_{r},0\big), ~~\text{and} ~~ \beta_n = \text{max}\big(L^0_{l},0\big)
\end{equation*}

As an example, \cref{eq:closed_form_exmp_ub,eq:closed_form_exmp_lb} are the closed form expressions for the upper bound (blue curve) and lower bound (red curve) linear piecewise over-approximations shown in \cref{fig:fig_1d}c.  

\begin{equation}\label{eq:closed_form_exmp_ub}
\begin{split}
    g_\text{ub}(x) =&  (\num{-0.9171523356672744}  \max(0, \num{-1.24381557588518}  (x - \num{-0.7668186154783817})) +  \\
    & \num{-0.6450757227359059}  \max(0, \min(\num{1.24381557588518}  (x - \num{-1.5707963267948966}), \num{-1.304089363266367}  (x - 0.0))) + \\
    & 0.0  \max(0, \min(\num{1.304089363266367}  (x - \num{-0.7668186154783817}), \num{-1.2598749193762386}  (x - \num{0.7937295874538862}))) + \\
    & \num{0.7937295874441845}  \max(0, \min(\num{1.2598749193762386}  (x - 0.0), \num{-1.2868907512989782}  (x - \num{1.5707963267948966}))) + \\
    & \num{0.9171523356672744}  \max(0, \num{1.2868907512989782}  (x - \num{0.7937295874538862})))
\end{split}
\end{equation}

\begin{equation}\label{eq:closed_form_exmp_lb}
\begin{split}
    g_\text{lb}(x) =& (\num{-0.9171523356672744}  \max(0, \num{-1.2868907512989782}  (x - \num{-0.7937295874538862})) + \\
    & \num{-0.7937295874441845}  \max(0, \min(\num{1.2868907512989782}  (x - \num{-1.5707963267948966}), \num{-1.2598749193762386}  (x - 0.0))) + \\
    & 0.0  \max(0, \min(\num{1.2598749193762386}  (x - \num{-0.7937295874538862}), \num{-1.304089363266367}  (x - \num{0.7668186154783817}))) + \\
    & \num{0.6450757227359059}  \max(0, \min(\num{1.304089363266367}  (x - 0.0), \num{-1.24381557588518}  (x - \num{1.5707963267948966}))) + \\
    & \num{0.9171523356672744}  \max(0, \num{1.24381557588518}  (x - \num{0.7668186154783817})))
\end{split}
\end{equation}

\subsection{Single Pendulum Example}
To illustrate the procedure for obtaining a relational overapproximation, we use the inverted single pendulum  dynamical system. 
We derive the discrete-time dynamical system from the governing differential equations. A \emph{simple} single pendulum is comprised of a point mass $m$ and a massless rod of length $\ell$. The pendulum is actuated by a motor that exerts a torque of $u$. The goal is to keep the pendulum in the up-right position. Neglecting air resistance, the governing differential equation for this dynamical system is:
\begin{equation}\label{eq:single_pend_eq_continuous}
    \ddot{\theta} = \frac{g}{\ell} \sin\theta + \frac{1}{m\ell^2}u
\end{equation}
where $\theta$ denotes the angle made by the pendulum arm and vertical direction and $g$ is the gravitational acceleration.
We use a first-order Euler scheme with a temporal step size of $\Delta \tau$ to produce the discrete-time equations:
\begin{subequations} \label{eq:single_pend_eq_1}
\begin{align*}
    x_{t+1, 1} = &x_{t, 1} + \Delta \tau ~ x_{t, 2} \\
    x_{t+1, 2} = &x_{t, 2} + \Delta \tau \left( \frac{g}{\ell} \sin x_{t, 1} + \frac{1}{m\ell^2} u_t \right)
\end{align*}
\end{subequations}
where $ \mathbf{x}_t = [ x_{t, 1}, x_{t, 2} ]^T = [ \theta_t, \dot\theta_t ]^T $ and $u_t$ represents the discrete state variable and control input, respectively, at time step $t$. 
In order to obtain the relational overapproximation of this system, we need to specify the domain of input parameters (states and control parameters). 
For the state parameters, the domain is typically specified by the user. 
The control signal $u_t$ would normally be specified by a neural network, and the range of $u_t$ for the dynamics approximation could be found using any method for estimating the output range of a neural network, such as
interval arithmetic, linear relaxation, or Lipschitz constant estimation of the network \citep{tjeng2017evaluating, xiang2018output}. 
However in this simple example, we simply specify a set that $u_t$ lies within. 
Assuming the initial set of states $(x_{0,1}, x_{0,2})$ lie within $[-1,1]^2$, and initial control policy input lies within $[-2, 2]$, 
 the relational overapproximation of this system that specifies states at the next time step ($t=1$) is given by: 
\begin{subequations} \label{eq:single_pend_eq_2}
\begin{align*}
    &x_{1, 1} = x_{0, 1} + \Delta \tau ~ x_{0, 1} \\
    &-1 \leq x_{0, 1} \leq  1 \\
    &-1 \leq x_{0, 2} \leq 1  \\
    &x_{1, 2} = x_{0, 2} + \Delta \tau \left( \frac{g}{\ell} v_1 + \frac{1}{m\ell^2} u_0 \right)\\
    &v_1 \leq e_{\text{UB}_{v_1}}(x_{0,1}) \\
    &v_1 \geq e_{\text{LB}_{v_1}}(x_{0,1}) \\
    &-2 \leq u_0 \leq 2 
\end{align*}
\end{subequations}
The two functions $e_{\text{LB}_{v_1}}$ and $e_{\text{UB}_{v_1}}$ are the lower and upper bounds of $\sin(x_{0,1})$, which can be computed using the \Call{GetUpperBound}{} and \Call{GetLowerBound}{} functions. For example, if we use piecewise linear functions with two linear segments, we obtain 
\begin{subequations} \label{eq:sin_abstraction}
\begin{align*}
    \begin{split}
    e_{\text{UB}_{v_1}}(x) = &- 1.00 \text{max}(0, -1.449(x +0.881)) \\&
           - 0.761\text{max}(0, \text{min}(1.449(x + 1.570), -1.135x)) \\ &
           + 0.01 \text{max}(0, \text{min}(1.135(x + 0.881), x - 1.0)) \\&
           + 1.01 \text{max}(0, \text{min}(x, -1.752(x - 1.571)) \\ &
           + 1.01 \text{max}(0, 1.752(x - 1.0)),
    \end{split}
          \\
    \begin{split}
    e_{\text{LB}_{v_1}}(x) =& - 1.01 \text{max}(0, -1.752(x + 1.0)) \\&
           - 1.01 \text{max}(0, \text{min}(1.752(x + 1.571), -x)  \\ &
           - 0.01 \text{max}(0, \text{min}(x + 1.0, -1.135(x - 0.881))) \\&
           + 0.761\text{max}(0, \text{min}(1.135x, -1.449(x - 1.571)) \\&
           + 0.99 \text{max}(0, 1.449(x - 0.881))
    \end{split}
\end{align*}
\end{subequations}

\subsection{Solving Reachability Problems} \label{sec:problemtypes}
OVERT solves reachability problems, meaning it reasons about the set of states reachable over time and whether this set of states intersects with an unsafe set or reaches a goal.
One way to frame the reachability problem is to explicitly compute the reachable set of the closed-loop system.
The \emph{reachable set} at a given timestep, $\mathcal R_t$, is comprised of all possible states that the system could visit at time $t$.
If the intersection of the reachable set and the unsafe set is empty,
$ \mathcal R_t \cap \mathcal S_{\text{unsafe}} = \varnothing $,
the safety property holds at time $t$. 
If the reachable set is a subset of the goal set, $\mathcal R_t \subseteq \mathcal S_{\text{goal}}$, the system is guaranteed to reach the goal at time $t$. 

Another way to frame the reachability problem is by solving what we call \emph{feasibility problems}.
Feasibility problems can directly encode the unsafe set (complement of the safe set) and return SAT indicating that the unsafe set is reachable or UNSAT indicating that the unsafe set is not reachable, without explicitly computing the reachable set. 

OVERT can solve reachability problems using either an explicit reachable set computation framing or a feasibility framing, and these two approaches are often complementary. 
Since reachable sets are typically overapproximated by simple geometric shapes such as  hyperrectangles or polytopes, reachable set computation introduces some looseness into the approximation. 
Conversely, solving feasibility problems incurs less looseness and less wrapping effect because the reachable set is instead represented implicitly. 
Consequently, solving feasibility problems results in fewer spurious counter examples and allows more properties to be proven.
However, computing reachable sets provides intuition about the the evolution of the system and allow the reachability problem to remain tractable over long time horizons.

We can compute reachable sets in two different ways.
The first approach is what we call the \emph{concrete} approach. 
Beginning with all state variables $\x$ constrained to lie in the input set $\mathcal I$ at time $t=0$, this approach computes a concrete reachable set $\mathcal R_1$ of the system at timestep $t=1$. 
For the next timestep, a new problem instance is created with the state variables $\x$ now constrained to lie within $\mathcal R_1$, and the concrete reachable set $\mathcal R_2$ is calculated, which is an overapproximation of the reachable set at timestep $t=2$. 
Here, we define the concrete reachable set as an explicit representation of reachable set (e.g., $1.56 \leq x \leq 4.67$).
The term \emph{concretizing} refers to computing such a concrete reachable set.
This approach effectively resets the reachable set computation problem, yielding a 1-step problem where only the initial set changes at each step. 
Concretizing after just one step incurs compounding looseness after several iterations, rendering the reachable set more conservative. 
This looseness allows for more spurious counter examples.
However, computing 1-step reachable sets is very fast. 

The second approach is what we call the \emph{symbolic} approach. In this approach, an implicit, symbolic representation of the reachable set is unrolled for several timesteps, and only concretized at the final timestep in the sequence. 
The symbolic approach produces sets that are significantly tighter, but the computational complexity of this approach can grow quickly. Each additional timestep encoded in the symbolic representation adds an approximately equal number of additional constraints. If one were to use only this approach, the length of time over which one could check properties would be severely constrained. 

In order to keep the problem computationally tractable, and yet obtain a reasonably accurate reachable set, we employ a hybrid approach, where concretization is performed only when the reachable set begins to get too large. We refer to this approach as the \emph{hybrid-symbolic} approach, which is implemented by the function $\Call{ComputeReachSets}{\vect{f}, \mathcal I, \text{cI}}$ (\cref{alg:hyb-sym}), where $\vect{f}$ is the closed-loop system, $\mathcal I$ is an input set, and cI 
 stands for \textit{concretization intervals} and is an array such as $[5,10,5]$ indicating that we concretize at timesteps $5,~15,~20$.
For example, we might compute nine one-step concrete sets, and then calculate a tight symbolic reachable set at timestep 10, beginning from a concrete set at timestep 1. This alternation is then repeated, calculating one-step concrete sets and symbolic sets when needed. 
The use of a hybrid-symbolic approach is one aspect that sets OVERT apart from related work by \citet{julian2021reachability} and \citet{tran2020nnv3}.
    \begin{algorithm}[H]
        \caption{Hybrid-Symbolic Reachable Set Computation}\label{alg:hyb-sym}
        \begin{algorithmic}[1]
        \Function{ComputeReachSets}{$\vect{f}$, $\mathcal I$, cI} 
        \State sym\_input\_set $\gets \mathcal I$ 
        \State $\mathcal R_i \gets \mathcal I$ 
        \State reach\_sets $\gets \emptyset$
        \State approximations $\gets \emptyset$
        \For{$n$ in cI}
            \For{$i \gets 1:n - 1$}
                \State $\hat{\vect{f}}_i \gets$ \Call{Overapproximate}{$\vect{f}$, $\mathcal R_i$}
                \State Add $\hat{\vect{f}}_i$ to approximations
                \State $\mathcal R_i \gets$ \Call{OneStepReach}{$\hat{\vect{f}}_i$,$\mathcal R_i$} 
                \State Add $\mathcal R_i$ to reach\_sets 
            \EndFor
            \State $\hat{\vect{f}}_i \gets$ \Call{Overapproximate}{$\vect{f}$, $\mathcal R_i$}
            \State Add $\hat{\vect{f}}_i$ to approximations
            \State $\mathcal R_n \gets$ \Call{SymbolicReach}{approximations, sym\_input\_set, $n$} 
            \State Add $\mathcal R_n$ to reach\_sets  
            \State $\mathcal R_i \gets \mathcal R_n$ 
            \State sym\_input\_set $\gets \mathcal R_n$
        \EndFor
        \State \Return reach\_sets
        \EndFunction
        \end{algorithmic}
    \end{algorithm}
    \begin{algorithm}
\begin{algorithmic}[1]
        \caption{Feasibility of Invariant Properties}\label{alg:feas_G}
        \Function{$\text{Feasibility}_G$}{$\vect{f}$, $\mathcal I$, $n$, $P$} 
        \State input\_set $\gets \mathcal I$  
        \State $\mathcal R_i \gets \mathcal I$ 
        \State approximations $\gets \emptyset$ 
        \State holds $\gets$ true  
        \State $i \gets 1$
        \While{$i \leq n$ and holds}
            \State $\hat{\vect{f}}_i \gets$ \Call{Overapproximate}{$\vect{f}$, $\mathcal R_i$} 
            \State Add $\hat{\vect{f}}_i$ to approximations
            \State holds $\gets$ holds and \Call{MultiStepFeas}{approximations, input\_set, $i$, $P$}  
            \State $\mathcal R_i \gets$ \Call{OneStepReach}{$\hat{\vect{f}}_i$, $\mathcal R_i$}
            \State $i \gets i + 1$    
        \EndWhile
        \State \Return holds
        \EndFunction
\end{algorithmic}
    \end{algorithm}
    
    \begin{algorithm}[H]
\begin{algorithmic}[1]
        \caption{Hybrid-Symbolic Feasibility}\label{alg:hyb-sym-feas}
        \Function{$\text{HSFeasibility}_G$}{$\vect{f}$, $\mathcal I$, $n$, $P$} 
        \State input\_set $\gets \mathcal I$  
        \State $\mathcal R_i \gets \mathcal I$ 
        \State approximations $\gets \emptyset$ 
        \State holds $\gets$ true  
        \State $i \gets 1$
        \While{$i \leq n$ and holds}
            \State $\hat{\vect{f}}_i \gets$ \Call{Overapproximate}{$\vect{f}$, $\mathcal R_i$} 
            \State Add $\hat{\vect{f}}_i$ to approximations
            \State holds $\gets$ holds and \Call{MultiStepFeas}{approximations, input\_set, $i$, $P$} 
            \If{time to reset}
                \State $\mathcal R_i \gets$ \Call{SymbolicReach}{approximations, input\_set, $i$}
                \State input\_set $\gets \mathcal R_i$
                \State $n \gets n - i$
                \State $i \gets 1$
            \Else
                \State $\mathcal R_i \gets$ \Call{OneStepReach}{$\hat{\vect{f}}_i$, $\mathcal R_i$}
                \State $i \gets i + 1$  
            \EndIf
        \EndWhile
        \State \Return holds
        \EndFunction
\end{algorithmic}
    \end{algorithm}
    
The function \Call{Overapproximate}{$\vect{f}$, $\mathcal R_i$} calculates an overapproximation of the closed-loop system $\vect{f}$ over the domain specified by $\mathcal R_i$.
The function \Call{OneStepReach}{$\hat{\vect{f}}_i$, $\mathcal R_i$} computes a one step reachable set for the system from $\mathcal R_i$, and the function 
\Call{SymbolicReach}{approximations, sym\_input\_set, $n$} computes a reachable set for the system $n$ steps into the future using symbolic relations between timesteps. 
The function \Call{OneStepReach}{} is defined \Call{OneStepReach}{$\hat{\vect{f}}_i$, $\mathcal R_i$} = \Call{SymbolicReach}{$\hat{\vect{f}}_i$, $\mathcal R_i$, $1$}.
More detailed on how the functions are implemented will be discussed in \cref{sec:solve-mip}.

OVERT solves feasibility problems by iteratively unrolling the closed-loop system and checking if the complement of the desired property is UNSAT at the final time. 
However, OVERT requires a domain for the state variables in order to overapproximate the dynamics at each subsequent timestep. 
OVERT provides this domain for feasibility problems by calculating 1-step reachable sets. 
This is illustrated in \cref{alg:feas_G}.
The function \Call{MultiStepFeas}{approximations, $\mathcal I$, $n$, $P$} performs a symbolic feasibility query to determine if property $P$ holds $n$ steps into the future beginning from set $\mathcal I$.
Feasibility for goal-reaching properties, or $F$ properties, would be written very similarly to \cref{alg:feas_G}, except it would be checking for the first instance of the property holding, instead of checking that it holds at every step.

In our experiments, the feasibility approach shown in \cref{alg:feas_G} ran fast enough that a hybrid-symbolic-feasibility approach was not necessary.
However, one such algorithm mixing the use of feasibility problems and symbolic reachable set computation for a property of the form $G~P$ (globally $P$) is described in \cref{alg:hyb-sym-feas}.
Such an approach would keep the problem tractable for long time histories.

\subsection{MIP Formulation \label{sec:solve-mip}}
Once the system dynamics have been abstracted, the closed-loop system consists of a conjunction of linear and piecewise linear constraints. 
The piecewise linear constraints used are $\min$, $\max$ and $\text{relu}$.
The key insight of several neural network verification tools for ReLU activations~\citep{akintunde2018reachability, tjeng2017evaluating} is that piecewise linear activation functions can be encoded into an optimization problem using binary variables, resulting in a mixed integer program (MIP). 

We compute reachable sets by constructing an optimization problem where the abstracted closed-loop system dynamics are encoded as constraints.
An initial set for the state variables is also encoded as a constraint.
We then find the minimum and maximum value of each state variable at subsequent timesteps, subject to these constraints. 
The symbolic reachability procedure, \Call{SymbolicReach}{}, solves the following two optimization problems for the $k$th state component in order to compute the reachable set $n$ steps in the future:
\begin{mini*}|l|
  {}{\x_{n,k}}{}{}
    \addConstraint{\x_{i}}{\in \mathcal S}{}
  \addConstraint{\vect x_{t+1}}{\bowtie \hat{\vect{f}} (\x_t, \vect c (\x_t)),}{~t=i,\ldots,n-1}
 \end{mini*}

\begin{maxi*}|l|
  {}{\x_{n,k}}{}{}
    \addConstraint{\x_{i}}{\in \mathcal S}{}
  \addConstraint{\vect x_{t+1}}{\bowtie \hat{\vect{f}} (\x_t, \vect c (\x_t)),}{~t=i,\ldots,n-1}
 \end{maxi*}
Here, timestep $i$ is the initial timestep and set $\mathcal S$ is the starting set. The symbol $\hat{\vect{f}}$ represents the overapproximated system dynamics, $\vect c$ represents the control policy, and $\bowtie$ represents that there is a relation between $\vect x_{t+1}$ and $\hat{\vect{f}} (\x_t, \vect c (\x_t))$. 
The minimum and maximum values specify the reachable set as a hyper-rectangle. This process is repeated for all components of the state. For example, to compute the reachable set $10$ steps in the future beginning from timestep $0$, we would set $i=0$, $\mathcal S = \mathcal I$, and $n=10$.
Once the reachable sets have been computed, they are then intersected with the unsafe sets or goal sets to determine if the desired property holds.

In order to ensure soundness in the finite-precision regime, 
the solution to the dual optimization problem, rather than the primal problem, is used to construct the overapproximate reachable set. 
For minimization problems, the dual solution is guaranteed to be a lower bound, and for maximization problems, the dual solution is guaranteed to be an upper bound. Thus using the dual bound increases the volume of the hyperrectangle and still represents an overapproximation of the true reachable set.
Most modern optimization software provides access to the dual as well as the primal solution.
In practice, the duality gap between the primal and dual solutions is small, and using the dual does not add significant looseness. 

In order to solve the feasiblity framing of reachability, a similar problem is constructed where the closed-loop system dynamics are encoded as constraints and an initial set is specified for the state variables.
In this case, however, no objective is specified and the negation of the desired property is encoded as an additional constraint. 
The symbolic feasibility procedure, \Call{MultiStepFeas}{}, solves the following feasibility problem:
\begin{mini*}|l|
  {}{0}{}{}
    \addConstraint{\x_{i}}{\in \mathcal S}{}
  \addConstraint{\vect x_{t+1}}{\bowtie \hat{\vect{f}} (\x_t, \vect c (\x_t)),}{~t=i,\ldots,n-1}
  \addConstraint{\neg P(\x_n)}{}{}
 \end{mini*}
 If no feasible solution can be found, the negation of the property is \emph{unsatisfiable} and the property is proven, meaning that the system, for example, never visits (at a given time $n$) the unsafe set.  
Unlike explicit reachable set computation, this approach does not need to solve optimality problems and therefore is typically less computationally expensive. 

There are many efficient tools for solving MIPs, which allows for tractable verification of neural networks. 
An important factor contributing to the speed of verification is the encoding used to represent the ReLUs and other piecewise linear activations in the MIP. 
In this work, a ReLU encoding inspired by the MIPVerify algorithm \citep{tjeng2017evaluating} is used. 
This encoding is a tighter alternative to the \emph{big-M} formulations that are commonly used for encoding min and max operations (\cite{akintunde2018reachability}). 
The resultant MIP is then solved using Gurobi \citep{gurobi}. 

The ReLU encoding presented in the MIPVerify algorithm requires a bound for the values of all neurons in the neural network. 
Such bounds can be computed using interval arithmetic or a linear programming relaxation \citep{tjeng2017evaluating}.
In this work, we use a Lipschitz-constant-based bounding algorithm designed by \citet{xiang2018output}.
An encoding inspired by MIPVerify is also used for the piecewise-linear $\min$ and $\max$ functions in the abstraction of the dynamics. 
This encoding is particularly suitable for OVERT, as bounds on each variable are already calculated during the abstraction of the dynamics, and these bounds can be used for the MIP encoding.
OVERT is currently implemented in Julia. We have open-sourced two Julia packages containing implementations of the sound overapproximation\footnote{\url{https://www.github.com/sisl/OVERT.jl}} and verification algorithms.\footnote{\url{https://www.github.com/sisl/OVERTVerify.jl}}

\section{Experiments} \label{sec:exp}

Below we demonstrate application of OVERT to a number of benchmark examples. All simulations are conducted with a machine with two 14-core Intel(R) Xeon(R) CPU E5-2690 v4 CPUs @ 2.60GHz (28 cores total), 128 GB RAM, and Ubuntu 18.04.

\subsection{Benchmark Examples} \label{sec:benchmark}
For the purpose of demonstration, we have chosen four classical control systems as described below.
These four examples were part of the ARCH-COMP AINNC 2020 competition \citep{tranarch}. We have also trained additional control policies to facilitate comparison. 


\begin{table}[]
\caption{Benchmark Problems.\label{table:problem_description}}
\begin{center}
\begin{footnotesize}
\begin{tabular}{@{}llrr@{}}
\toprule
Problem & Dynamics        & Neural Network  Controller Size              & Number of Timesteps                                                             \\ \midrule
\textbf{S1}                    & single pendulum & $2\times 25 \times 25 \times 1$              & 25                                           \\
\textbf{S2}                    & single pendulum & $2\times 50 \times 50 \times 1$              & 25                                            \\
\textbf{T1}                    & tora            & $4\times 25 \times 25 \times 25 \times 1$ & 15 \\
\textbf{T2}                    & tora            & $4\times 50 \times 50 \times 50 \times 1$    & 15 \\
\textbf{T3}                    & tora            & $4\times 100 \times 100 \times 100 \times 1$    & 15 \\
\textbf{C1}                    & car             & $4\times 100 \times 1$                       & 10 \\
\textbf{C2}                    & car             & $4\times 200 \times 1$                       & 10 \\
\textbf{C3}                    & car             & $4\times 300 \times 1$                       & 10 \\
\textbf{ACC}                   & adaptive cruise control  & $6\times 3\times 20\times 20\times 20\times 1$ & 55 \\
\bottomrule
\end{tabular}
\end{footnotesize}
\end{center}
\end{table}

\begin{enumerate}
    \item \textbf{Single Pendulum (S)}: The governing equation for an inverted single pendulum is \cref{eq:single_pend_eq_continuous}. 
    We will use the discrete-time version shown in \cref{eq:single_pend_eq_repeat} with state variables $ \mathbf{x}_t = [ x_{t, 1}, x_{t, 2}]^T = \left[ \theta_t, \dot\theta_t \right]^T $ and input control torque $u_t$ which is repeated below for the sake of completeness: 
    \begin{subequations} \label{eq:single_pend_eq_repeat}
    \begin{align}
        x_{t+1, 1} = &x_{t, 1} + \Delta \tau ~ x_{t, 2} \\
        x_{t+1, 2} = &x_{t, 2} + \Delta \tau \left( \frac{g}{\ell} \sin x_{t, 1} + \frac{1}{m\ell^2} u_t \right)
    \end{align}
    \end{subequations}
  
    The parameter values are $m=0.5$, $\ell=0.5$, $g=1.0$ and $\Delta \tau=0.1$. The state space $[x_1, x_2]$ is two-dimensional, and the control input $u$ is one-dimensional.
    
    Control policies are trained to stabilize the pendulum upside-down using behavior cloning, a supervised learning approach for training control policies. Here, a neural network is trained to replicate expert demonstrations. We initially generate a set of expert control inputs for trajectories originating from different initial states of the system. Expert control inputs are defined as those leading the system to reach to its goal state, $0$ radians, in finite time. The expert control inputs are generated using optimal control techniques. Specifically, we have used an implementation of the LQR (Linear Quadratic Regulator) algorithm to generate expert control inputs.
    Two control policies with different architectures were trained using this data; details are given in \cref{table:problem_description}.  \\
    \textbf{Property:} The angle of the pendulum must stay above $\approx -15$ degrees, or $-0.2167$ radians, for 25 discrete time steps: 
    \begin{equation*}
    G_{1:25}~x_{1} \geq -0.2167    
    \end{equation*}
    beginning from the initial set $$x_{1}, x_{2} \in [1, 1.2]\times[0, 0.2]$$
   
    \item \textbf{TORA (T)}: This example is a model for rotational actuators. TORA stands for Transalation Oscillations of a Rotational Actuator. The system consists of a cart that can roll along a frictionless surface and is attached to the wall by a spring. Inside the cart, there is a mass on a rod that can rotate to actuate the cart. The state may be described by the x-displacement of the cart and the angular displacement of the rotational actuator. 
    However, \citet{jankovic1996tora} derive the following simpler equations of motion using more complex state variables:
    \begin{subequations}
        \begin{align*}
        \dot{x}_1 =& x_2 \\
        \dot{x}_2 =& -x_1 + \epsilon\sin x_3\\
        \dot{x}_3 =& x_4\\
        \dot{x}_4 =& u
        \end{align*}
    \end{subequations}
    The model has a four-dimensional state space $[x_1, x_2, x_3, x_4]$ and one-dimensional control input $u$. We will use the following discrete form of the model:
    \begin{subequations}
        \begin{align*}
            x_{t+1, 1} =& x_{t, 1} + \Delta \tau x_{t, 2}\\
            x_{t+1, 2} =& x_{t, 2} + \Delta \tau \left(\epsilon\sin x_{t,3} -x_{t, 1} \right) \\
            x_{t+1, 3} =& x_{t, 3} + \Delta \tau x_{t, 4}\\
            x_{t+1, 4} =& x_{t, 4} + \Delta \tau u_t
        \end{align*}
    \end{subequations}
    The parameter values are $\epsilon = 0.1$ and $\Delta \tau = 0.1$.
    We tested this benchmark with three different neural network control policies, as shown in \cref{table:problem_description}. The largest neural network is that of \citet{dutta2019reachability}. This network is trained to stabilize the cart at $x=0$. We prepared the two smaller networks using supervised learning on data generated from the larger network. This way, we ensure that the three networks are trained to do a similar task, and therefore the difference in computational time is due to the different neural network architecture.\\ 
    \textbf{Property:} The first state variable, $x_{1}$ must stay close to the origin:
    \begin{equation*}
    G_{1:15}~x_{1} \in [-2,2]
    \end{equation*}
    beginning from the initial set $$x_1, x_2, x_3, x_4 \in [0.6, 0.7] \times [-0.7,-0.6] \times [-0.4, -0.3] \times [0.5, 0.6]$$
    
    \item \textbf{Car (C)}: This example uses a kinematic bicycle model to approximate car dynamics \citep{rajamani2011vehicle, kong2015kinematic}.
    The state of the system may be described by the position of the car ($x_1, x_2$) as well as the yaw angle ($x_3$) and speed ($x_4$).
    The ordinary differential equations governing motion are: 
    \begin{subequations}
        \begin{align*}
            \dot x_1 =& x_4\cos \left(x_3\right)\\
            \dot x_2 =& x_4\sin \left(x_3\right)\\
            \dot x_3 =& u_2\\
            \dot x_4 =& u_1
        \end{align*}
    \end{subequations}
    The state space $[x_1, x_2, x_3, x_4]$ is four-dimensional, and control inputs $[u_1, u_2]$ are two-dimensional.  
    We will use the following discrete form of the model:
    \begin{subequations}
        \begin{align*}
            x_{t+1, 1} =& x_{t, 1} + \Delta \tau x_{t, 4}\cos \left(x_{t,3}\right)\\
            x_{t+1, 2} =& x_{t, 2} + \Delta \tau x_{t, 4}\sin \left(x_{t,3}\right)\\
            x_{t+1, 3} =& x_{t, 3} + \Delta \tau u_{t, 2}\\
            x_{t+1, 4} =& x_{t, 4} + \Delta \tau u_{t,1}
        \end{align*}
    \end{subequations}
    We set
    $\Delta \tau =0.2$.
    
    Similar to the Tora benchmark, we tested this example with three different neural network control policies. The largest neural network is that of \citet{dutta2019reachability} and we prepared the two smaller networks using supervised learning. \\
    \textbf{Property:} The position of the car will reach a goal set near the origin within 10 timesteps:
    \begin{equation*}
      F_{1:10}~(x_{1}, x_{2}) \in [-0.6, 0.6]\times[-0.2,0.2]  
    \end{equation*}
    beginning from the initial set
    \begin{equation*}
      x_1,x_2,x_3,x_4 \in [9.5,9.55]\times [-4.5, -4.45]\times [2.1, 2.11] \times [1.5, 1.51] 
    \end{equation*}

    \item \textbf{Adaptive Cruise Control (ACC)}:
    This system models an adaptive cruise control policy for a car on a highway. The system has a target velocity, but there is a lead vehicle ahead of the ego vehicle, and the ego must maintain a safe distance from the lead vehicle. The neural network control policy adjusts the longitudinal acceleration of the ego vehicle. 
    The position, velocity and acceleration of the lead vehicle are $x_1$, $x_2$, and $x_3$. The position, velocity, and acceleration of the ego vehicle are $x_4$, $x_5$, and $x_6$. 
    The continuous-time dynamics of the system \citep{tranarch} are:
    \begin{subequations}
        \begin{align*}
            \dot x_1 =& x_2\\
            \dot x_2 =& x_3\\
            \dot x_3 =& -2x_3 + 2a - 2\mu x_2^2\\
            \dot x_4 =& x_5\\
            \dot x_5 =& x_6\\
            \dot x_6 =& -2x_6 + 2u - 2\mu x_5^2
        \end{align*}
    \end{subequations}
    where $[x_1,x_2, x_3, x_4, x_5, x_6]$ is the 6-dimensional state space and $u$ is the single control input (acceleration of the ego car). Here, $a$ and $\mu$ are the lead car acceleration and friction factor, respectively. We use the following discrete form of the system:
    \begin{subequations}
        \begin{align*}
            x_{t+1,1} =& x_{t, 1} + \Delta \tau x_{t,2}\\
            x_{t+1,2} =& x_{t, 2} + \Delta \tau x_{t,3}\\
            x_{t+1,3} =& x_{t, 3} + \Delta \tau \left(-2x_{t,3} + 2a - 2\mu x_{t,2}^2\right)\\
            x_{t+1,4} =& x_{t, 4} + \Delta \tau x_{t,5}\\
            x_{t+1,5} =& x_{t, 5} + \Delta \tau x_{t,6}\\
            x_{t+1,6} =& x_{t, 6} + \Delta \tau \left(-2x_{t,6} + 2u_t - 2\mu x_{t,5}^2\right)
        \end{align*}
    \end{subequations}
    Similar to \citet{tranarch}, we choose $a=-2$, which models the lead vehicle decelerating, and a friction parameter $\mu=10^{-4}$. The discretization timestep is set to  $\Delta \tau =0.1$. \\
    \textbf{Property:} A safe distance between the ego vehicle and the lead vehicle must be maintained:
    \begin{align}
    G_{1:55}~x_{lead} - x_{ego} &\geq D_{safe}
    \label{eq:acc_property}
    \end{align}
    where $D_{safe} = T_{gap}*v_{ego} + D_{min}$, $x_{lead} = x_{1}$, $x_{ego} = x_{4}$, $v_{ego} = x_{5}$, $D_{min} = 10$, and $T_{gap} = 1.4$.
    The initial set is
    $$x_1, x_2, x_3, x_4, x_5, x_6 \in [90, 91]\times [10, 11]\times [30, 30.2] \times [30, 30.2] \times [0, 0.01] \times[0, 0.01]$$
\end{enumerate} 

Each benchmark example contains nonlinearities that are overapproximated.
One may wonder what the right choice for $n$, the number of linear segments within each region of constant convexity, is for each benchmark. 
On one hand, as $n$ increases, the overapproximation gets tighter. 
On the other hand, this overapproximation requires more binary variables in the mixed-integer representation.
Interestingly, we experimentally observe that increasing the number of binary variables by increasing $n$ does not immediately make the MIP problem harder to solve. 
In other words, in our experiments, the MIP program was significantly more sensitive to number of binary variables originating from the neural network, and less so to those originating from the abstraction of the dynamics. 
In all of our experiments, we chose $n$ large enough such that the maximum relative error between $f(x)$ and $g(x)$ is less than 2\%. 

\subsection{Comparison with other tools}\label{sec:compare}
Here, we will compare the performance of OVERT with two other tools that can be used for closed-loop verification of discrete-time problems. It is imperative to note that discrete-time problems are inherently different than continuous-time problems. Even if a discrete-time problem is obtained by discretizing a continuous system, these two systems are fundamentally different, and not directly comparable. 
Therefore, it is not appropriate to compare OVERT to tools designed for continuous-time problems such as Sherlock \citep{dutta2019reachability} or Verisig \citep{ivanov2018verisig}.

As mentioned earlier, OVERT can solve a reachability problem by framing it as either a direct reachable set computation or as a feasibility problem. To show its performance on feasibility problems, we compare it with dReal \citep{gao2013dreal}, which is an automated reasoning tool designed for solving nonlinear decision problems over the real numbers. As dReal is not designed to perform model checking for transition systems, a wrapper was written to translate the bounded time model checking problem that OVERT solves into the SMT2 format that is readable by dReal.
For direct reachable set computation, we compare OVERT with the Neural Network Verification (NNV) tool \citep{tran2020nnv3}, which is designed for computing reachable sets of closed-loop systems with a neural network control policy. We emphasize here that at the time of writing, NNV was designed for use only with continuous-time systems, and we had to implement support for discrete-time systems. The approximate star-based method was used within NNV.

The results of the comparisons are shown in \cref{table:comparison} on the S1 and T1 problems. For each example, we have assumed a 12-hour timeout. 
For the reachable set computation framing of the problem, NNV finished the S1 problem in 483 seconds but was unable to prove the desired property as the approximate reachable set computed was too loose. In comparison, OVERT took just 99 seconds and was able to prove the property. For the S1 problem, we perform a symbolic query with OVERT at timesteps 10, 20, and 25, and propagate one-step concrete sets in between those timesteps.
On the T1 problem, NNV did not finish computation within 12 hours, but OVERT finished the task in less than 4 minutes and was able to prove the property. 
For the T1 problem, we perform a symbolic query with OVERT at timesteps 5, 10 and 15. 
It was noted that by timestep 8 in problem T1, NNV was tracking the reachable set of nearly 2 million polytopes, due to disjunctions introduced by the ReLU activations; this explosion in polytopes likely caused the timeout. 

Next, we compare OVERT to dReal for solving the feasibility framing of the problem.
dReal times out on the S1 problem, while OVERT takes less than two minutes to solve the feasibility framing of the problem. 
For the T1 problem, dReal is able to show that the property holds, but it takes approximately 3.5 hours, while OVERT takes less than two minutes to solve the feasibility framing. 
For both the S1 and T1 feasibility framing versions of the problem, \cref{alg:feas_G} was used, where no symbolic queries are performed to ``reset'' the time horizon.
If a symbolic query had been used, making the resultant feasibility problem e.g. only 10 steps long, the feasibility results would likely be even faster. This remains to be explored more fully in future work. 
The results clearly indicate that general purpose nonlinear reasoning tools such as dReal do not scale to handle neural networks, and specialized tools such as OVERT are necessary for the verification of systems containing neural networks. 

\begin{figure}[t]
\centering
\input{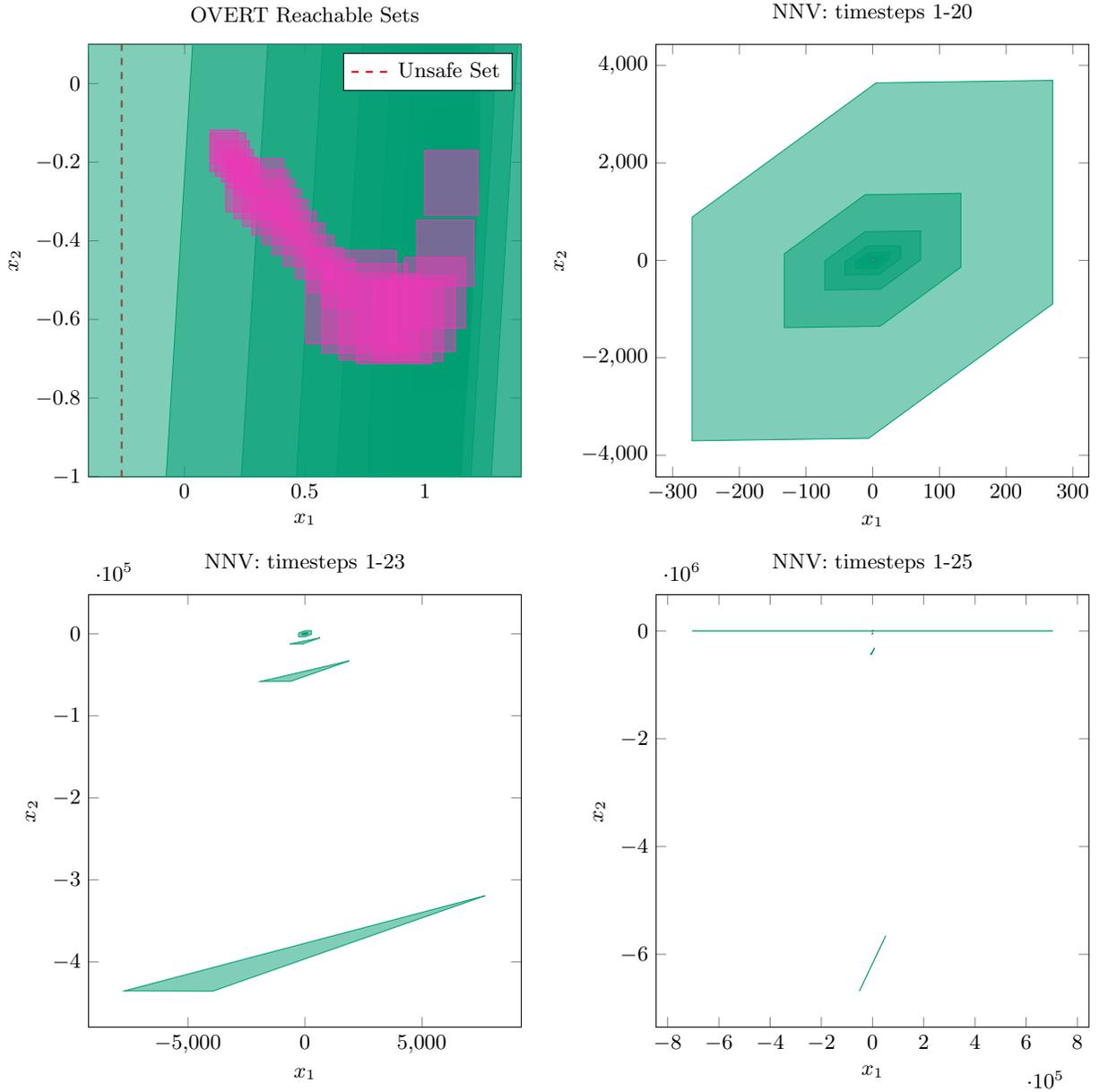}
\caption{\label{fig:nnv_comparison}Comparison of reachable set for the S1 problem using NNV (green polygons) and OVERT (pink polygons). The dashed red line in the top left plot shows the boundary of the unsafe set. }
\end{figure}

OVERT outperforms both dReal and NNV in terms of computation time, and the final reachable set obtained by OVERT are also significantly tighter than those found by NNV. \Cref{fig:nnv_comparison} shows the reachable sets obtained by OVERT and NNV for the S1 problem. As shown by \citet{xiang2018specification}, splitting the input set can produce tighter reachable sets. This phenomenon benefits the performance of both NNV and OVERT. To allow for fair comparison, we do not split the input set for either tool. As can be seen by comparing the top two plots in \cref{fig:nnv_comparison}, for the first 20 timesteps, the dimensions of the sets produced by NNV are up to 3 to 4 orders of magnitude larger than those produced by OVERT. For timesteps 21 to 25, however, shown in the bottom two plots, the sets appear to be degenerate as they cease to contain the true reachable set, which is still roughly centered near (0,0), and reduce to lines in the case of timesteps 24 and 25. As the sets for timesteps 21 to 25 grow very large with dimensions on the order of $10^6$, it is likely that numerical errors are creating the observed degeneracy. 
\begin{table}[]
\caption{Comparison between OVERT, NNV, and dReal \label{table:comparison}}
\centering
\resizebox{\textwidth}{!}{

\begin{tabular}{@{}lrrrrr@{}}
\toprule
Problem  & Timesteps   & DReal (s) & NNV (s) & $\text{OVERT}_\text{reach}$ (s) & $\text{OVERT}_\text{feas}$ (s) \\ \midrule
\textbf{S1} & 25            & Time Out, N/A & 483, fails     & 99, HOLDS            & \textbf{77}, HOLDS \\
\textbf{T1} & 15            & 12655, HOLDS  &  Time Out, N/A       & 232, HOLDS          & \textbf{79}, HOLDS \\ \bottomrule
\end{tabular}
}
 \end{table}

\subsection{Benchmarking Performance}\label{sec:results}
In this section, we report the results of reachability problems analyzed with OVERT. As aforementioned, each reachability problem can be posed either as a reachable set computation and then set intersection problem, or by directly encoding the property and solving a feasibility problem. We explore each of OVERT's two capabilities in the following two subsections. 

\subsubsection{Reachable Set Computation Problems}
\begin{figure}
    \centering
    \input{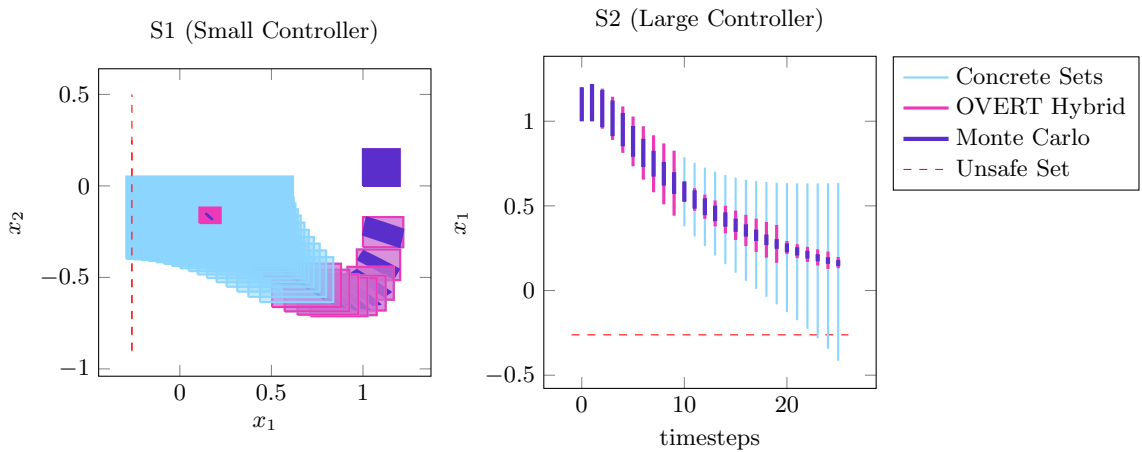}
    \caption{Reachability problem for the single pendulum benchmarks S1 and S2. The OVERT hybrid symbolic approach is compared to a naive approach where 1-step concrete sets are computed at every timestep as well as to the convex hull of of one million Monte Carlo simulations.  \label{fig:single_pendulum_reachability_controller}}
\end{figure}

We start with the single pendulum benchmark.
Fig.~\ref{fig:single_pendulum_reachability_controller} illustrates the reachability results for the single pendulum benchmark with two different neural network control policies. 
For both S1 and S2, symbolic reachable sets were computed at timesteps 10, 20, and 25.
It can be observed for both S1 and S2 that the hybrid-symbolic approach is capable of proving the safety property, but the naive 1-step concrete sets are too loose, and cannot prove the property.
The convex hull of Monte Carlo simulations represent an underapproximation of the true reachable set, and one can observe that the sets generated by OVERT's hybrid approach hug the true system trajectories tightly. 
This suggests that few spurious counter examples will be generated.

While one might observe that performing more and longer symbolic queries would produce even tighter reachable sets, this also adds to the complexity of the MIP to be solved. 
For example, in the case of S2, the naive one-step reachable set computation took less than 4 minutes, while the hybrid symbolic approach with two 10-step queries, one 5-step query, and 22 one-step queries required just under 12 minutes to solve.

When the update function of a dynamical system includes more nonlinear components, the complexity of the problem increases. Each nonlinear component requires piecewise linear overapproximation, which in turn translates into an MIP formulation with more binary variables.  While there is some variability from one problem to another, 
additional binary variables generally add complexity to the problem. 
Furthermore, a dynamical system with more nonlinearity may require a more complex control policy; e.g. a deeper neural network, which further increases the complexity of the problem.
In order to handle this additional complexity, in the Tora and Car examples, we reduced the number of timesteps and increased the frequency at which symbolic queries were performed to reset the time horizon. 

\begin{figure}
\begin{center}
    \input{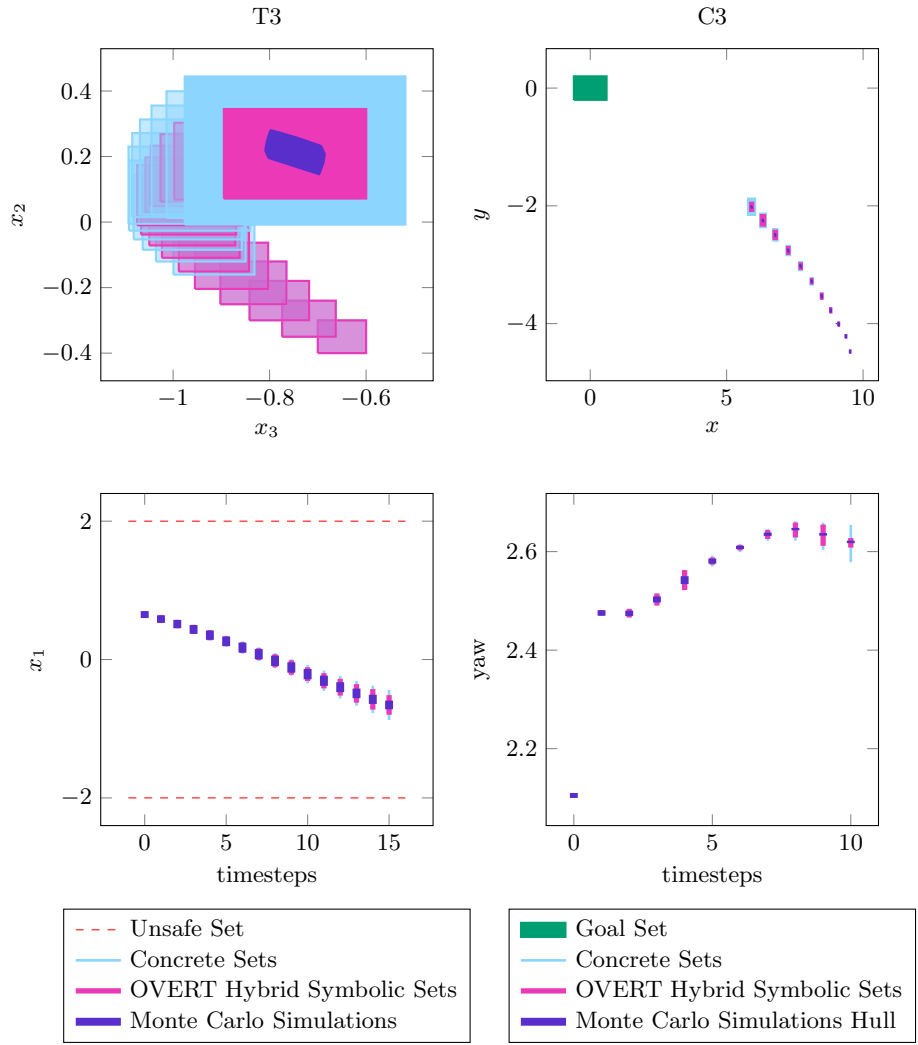}  
\end{center}
    \caption{Reachable sets for benchmark examples T3 (left) and C3 (right).
    Light blue indicates the reachable sets obtained via 1-step concretization and pink indicates those found with hybrid symbolic computation. The convex hull of 1 million Monte Carlo simulations of the original system is shown in purple. 
    \label{fig:car_tora}}
\end{figure}

\Cref{fig:car_tora} shows the reachability results for the Tora T3 example and Car C3 example. The difference between the concrete and hybrid-symbolic approaches is not as significant as in the single pendulum example, in part because we had to concretize the symbolic queries earlier to avoid having queries that were too large. Symbolic sets were calculated for all of the TORA problems (T1, T2, T3) at timesteps 5, 10, and 15. For the car problems (C1, C2, C3), symbolic sets were calculated at timesteps 5 and 10.  
The top left plot shows the 2-dimensional reachable sets for variables $x_{t,2}$ and $x_{t,3}$ for the T3 benchmark, and the bottom left plot show the 1-dimensional reachable sets for variable $x_{t,1}$ for the T3 benchmark. The reachable sets do not leave the safe set, indicated with dashed red lines, and so the safety property is proven.

The plots on the right side of \cref{fig:car_tora} show the reachability results for example C3. On the top right, the reachable sets of states $x_{t,1}$ and $x_{t,2}$ are shown. 
We observe that the reachable state sets are never a subset of the goal set indicating that we cannot prove the goal-reaching property holds, but we can show that the intersection of the reachable state sets and the goal set is empty, meaning we can prove that the goal-reaching property never holds. If the sets partially overlapped, the results would be inconclusive. 
This problem illustrates the versatility of computing the reachable state set concretely. Once the sets have been computed, multiple queries can be answered with minimal additional computation. In contrast, a whole new feasibility problem would have to be solved to address a second query.

\begin{figure}
    \centering
    \subcaptionbox{Measurement sets. It can be observed that the property holds, and that the ego vehicle does not get too close to the lead vehicle.\label{fig:acc_a}}{\input{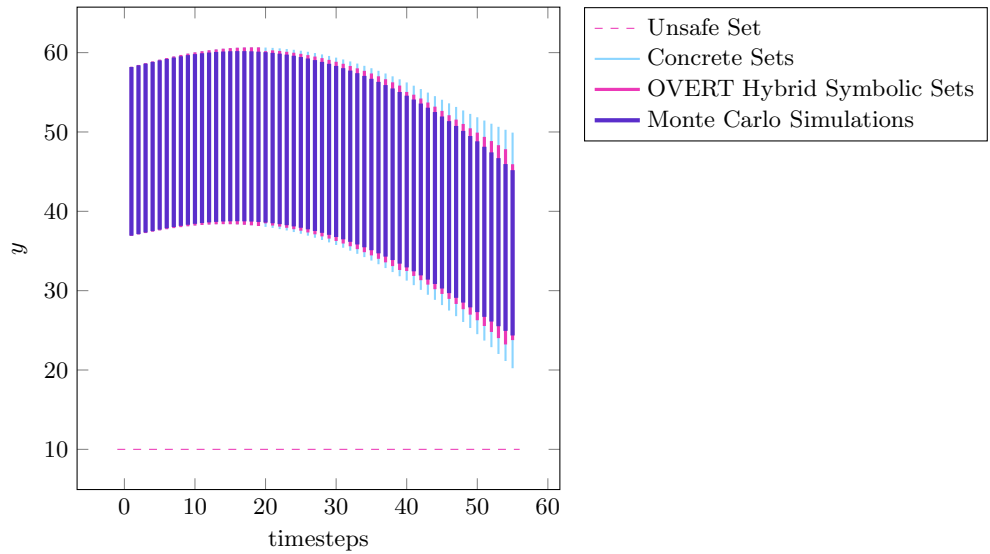}}\\[1cm]
    \subcaptionbox{State sets. A top-down view of the reachable sets of the ego car position, lead car position, and safe distance, which is based on ego velocity.\label{fig:acc_b}}{\input{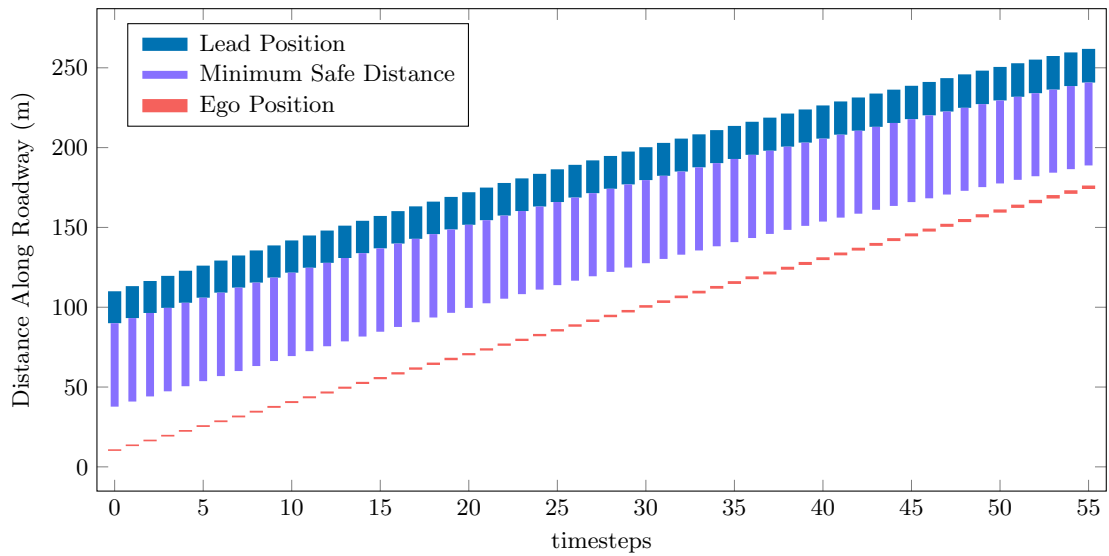}}
    \caption{Reachable sets for the Adaptive Cruise Control (ACC) benchmark example.}
    \label{fig:acc}
\end{figure}

\begin{figure}
    \centering
    \input{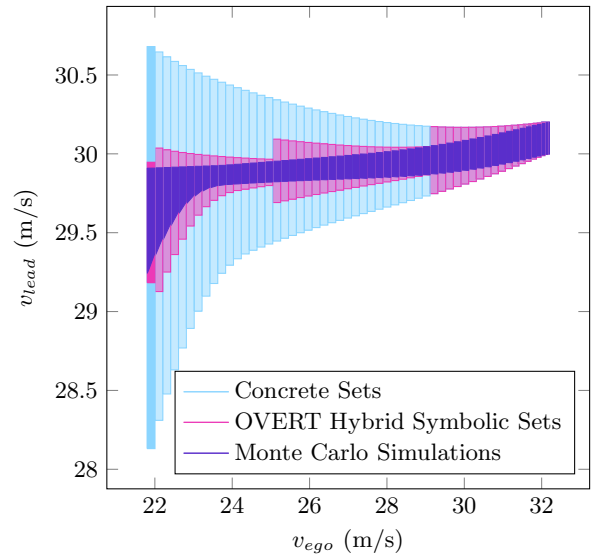}
    \caption{Reachable sets of velocity for the lead and ego vehicles for the Adaptive Cruise Control (ACC) benchmark example. Solid sets indicate the final timestep.}
    \label{fig:acc_2}
\end{figure}

The final example is the Adaptive Cruise Control benchmark which is shown in \cref{fig:acc,fig:acc_2}. \Cref{fig:acc_a} shows reachable sets of the measurement: 
\begin{align}\label{eq:acc_meas}
    y &= x_{lead} - x_{ego} - T_{gap} v_{ego}
\end{align}
over 55 timesteps, where symbolic queries were performed at timesteps 20, 40 and 55. \Cref{eq:acc_meas} is simply a re-arranged version of \cref{eq:acc_property}. In order to maintain safety, we must have that $y > D_{min} = 10$. 
One can see from the figure that this is indeed the case; the reachable sets do not cross the red dotted line indicating the edge of the unsafe set. 
One may also observe that the hybrid-symbolic sets offer modest tightness improvements over the concrete sets. This is due to inherent properties of the closed-loop system.
\Cref{fig:acc_b} shows a top-down view of the ego and lead car positions, as well as the $D_{safe}$, computed using OVERT's hybrid symbolic approach, and is intended to help the reader visualize the roadway scenario. In this depiction, the property could still hold even if there were overlap between the ego vehicle position and the safe distance threshold, as the ego vehicle maximum forward position, maximum safe distance, and lead car minimum forward position may not occur simultaneously.
Finally, \cref{fig:acc_2} shows 2-dimensional state sets of the ego vehicle and lead vehicle velocities. This plot demonstrates how tightly the OVERT hybrid-symbolic approach can hug the true system trajectories, compared to the looseness of naively computing 1-step reachable sets at every timestep.

\begin{figure}
    \centering
    \tiny
\begin{tikzpicture}[]
\begin{groupplot}[group style={horizontal sep=2cm, vertical sep=1.75cm, group size=2 by 1}]

\nextgroupplot [
  legend style = {at={(.5, 1.0)} , anchor=north west},
  ylabel = {Time (s)},
  title = {a) Reachable Set Computation Time},
  ymode = {log},
  xlabel = {Neurons},
  height=12cm, axis equal image, xmode=log, yminorticks=true, xminorticks=true, xtickten={1,2,3}, xmax=10^3, ymin=60, ymax=10^5, ymajorgrids=true, yminorgrids=true,
  xmode = {log}
]

\addplot+[
  solid, thick, blue,mark=none
] coordinates {
  (50.0, 99.0)
  (100.0, 682.0)
};

\addplot+[
  densely dotted, thick, symbolic_color, mark=none, mark options={fill=white},mark=none
] coordinates {
  (102.0, 409.0)
  (202.0, 517.0)
  (502.0, 896.0)
};

\addplot+[
  dashed, thick, red, mark options={fill=white},mark=none
] coordinates {
  (76.0, 232.0)
  (151.0, 3423.0)
  (301.0, 66987.0)
};

\addplot+[
  solid, thick, orange, mark options={fill=white},mark=none
] coordinates {
  (64.0, 846.0)
};

\addplot+[
  scatter,
  scatter src = explicit symbolic,
  only marks = {true},
  scatter/classes = {1={mark=*, blue}, 0={mark=square, blue}}
] coordinates {
  (50.0, 99.0) [1.0]
  (100.0, 682.0) [1.0]
};

\addplot+[
  scatter,
  scatter src = explicit symbolic,
  only marks = {true},
  scatter/classes = {1={mark=*, symbolic_color}, 0={mark=square, symbolic_color}}
] coordinates {
  (102.0, 409.0) [0.0]
  (202.0, 517.0) [0.0]
  (502.0, 896.0) [0.0]
};

\addplot+[
  scatter,
  scatter src = explicit symbolic,
  only marks = {true},
  scatter/classes = {1={mark=*, red}, 0={mark=square, red}}
] coordinates {
  (76.0, 232.0) [1.0]
  (151.0, 3423.0) [1.0]
  (301.0, 66987.0) [1.0]
};

\addplot+[
  scatter,
  scatter src = explicit symbolic,
  only marks = {true},
  scatter/classes = {1={mark=*, orange}, 0={mark=square, orange}}
] coordinates {
  (64.0, 846.0) [1.0]
};

\nextgroupplot [
  legend style = {at={(1.0, 1.0)} , anchor=north east},
  ylabel = {Time (s)},
  title = {b) Feasibility Computation Time},
  ymode = {log},
  xlabel = {Neurons},
  height=12cm, axis equal image, xmode=log, yminorticks=true, xminorticks=true, xtickten={1,2,3}, xmax=10^3, ymin=60, ymax=10^5, ymajorgrids=true, yminorgrids=true,
  xmode = {log}
]

\addplot+[
  solid, thick, blue,mark=none
] coordinates {
  (50.0, 77.0)
  (100.0, 558.0)
};
\addlegendentry{{}{Single Pendulum}}

\addplot+[
  densely dotted, thick, symbolic_color, mark=none, mark options={fill=white},mark=none
] coordinates {
  (102.0, 253.0)
  (202.0, 288.0)
  (502.0, 908.0)
};
\addlegendentry{{}{Car}}

\addplot+[
  dashed, thick, red, mark options={fill=white},mark=none
] coordinates {
  (76.0, 79.0)
  (151.0, 110.0)
  (301.0, 510.0)
};
\addlegendentry{{}{TORA}}

\addplot+[
  solid, thick, orange, mark options={fill=white},mark=none
] coordinates {
  (64.0, 625.0)
};
\addlegendentry{{}{Adaptive Cruise Control}}

\addplot+[
  scatter,
  scatter src = explicit symbolic,
  only marks = {true},
  scatter/classes = {1={mark=*, blue}, 0={mark=square, blue}}
] coordinates {
  (50.0, 77.0) [1.0]
  (100.0, 558.0) [1.0]
};
\addlegendentry{{}{Holds}}

\addplot+[
  scatter,
  scatter src = explicit symbolic,
  only marks = {true},
  scatter/classes = {1={mark=*, symbolic_color}, 0={mark=square, symbolic_color}}
] coordinates {
  (102.0, 253.0) [0.0]
  (202.0, 288.0) [0.0]
  (502.0, 908.0) [0.0]
};
\addlegendentry{{}{Does Not Hold}}

\addplot+[
  scatter,
  scatter src = explicit symbolic,
  only marks = {true},
  scatter/classes = {1={mark=*, red}, 0={mark=square, red}}
] coordinates {
  (76.0, 79.0) [1.0]
  (151.0, 110.0) [1.0]
  (301.0, 510.0) [1.0]
};

\addplot+[
  scatter,
  scatter src = explicit symbolic,
  only marks = {true},
  scatter/classes = {1={mark=*, orange}, 0={mark=square, orange}}
] coordinates {
  (64.0, 625.0) [1.0]
};

\end{groupplot}

\end{tikzpicture}
    \caption{Computation time versus number of ReLU neurons in the neural network control policy.}
    \label{fig:time}
\end{figure}
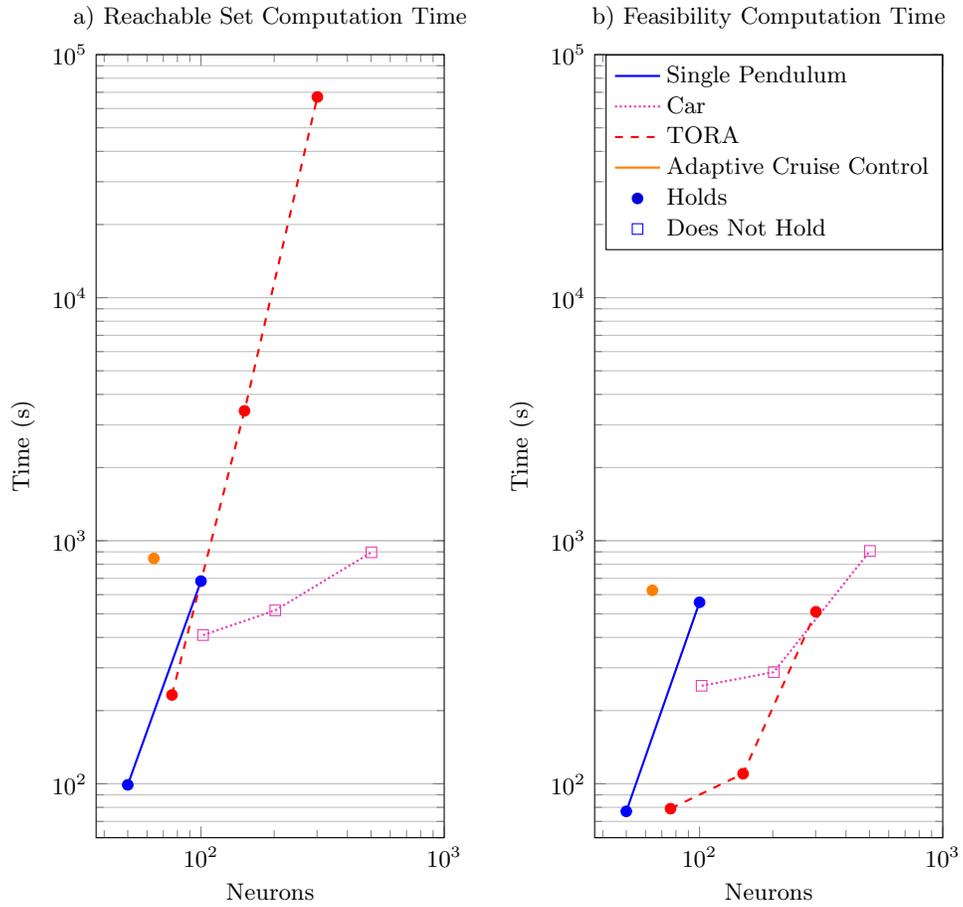

The compute times for each of the reachability problems are shown in \cref{fig:time}a. 
For smaller problems such as S1 and T1, the compute time is less than 5 minutes. For S2, C1, C2, C3, and ACC, the compute time is less than 15 minutes. However for the larger Tora examples, T2 and T3, 
the computation time for the reachable set computation framing begins to grow to the order of hours, with T3 taking a little over 18 hours.
While the Tora problems have control policies with similar numbers of ReLU activations as the other problems' control policies, the number of weights in the Tora problem control policies is larger than that of the other problems' control policies~(\cref{table:problem_description}). 

Additionally, note that the experimental timing results do not display a sharp trend with state dimension, as S2 is 2-dimensional problem, and ACC is a 6 dimensional problem with more than twice as many timesteps, yet ACC requires only 1.24 times as long to solve. This is as expected, as OVERT's reachable set computation has only explicit linear dependence on the state dimension (the number of optimization problems to be solved is $2n$, where $n$ is the state dimension). In the worst case, the number of activation regions in a network could be exponential in the input dimension (state dimension) \citep{katz2017reluplex}, which could make higher dimensional problems much slower, but our results do not exhibit this worst-case trend.

\subsubsection{Feasibility problems}
The next set of experiments explore solving the reachability problems described in section~\ref{sec:benchmark} as feasibility problems.
In addition to the closed-loop system and the initial sets for the input parameters, a feasibility query is comprised of a property. Specifically, the complement of the property that we would like to hold is encoded. 
In the case of the single pendulum for example, the property is staying within the safe set of $x_1 \geq -0.2167$ radians. The unsafe set, $x_1 \leq -0.2167$ radians, is encoded in the feasibility problem. 

A feasibility query in OVERT has two possible outcomes. 
If the solver returns UNSAT, the unsafe set cannot be reached and $x_1 \geq -0.2167$ for the specified time range. If the solver returns SAT, the pendulum may visit a state with angle $\theta \leq -0.2167$ and that the property cannot be guaranteed to hold. 
In this case, OVERT returns a counter example, which is a trajectory of the system.
As OVERT is based on overapproximating the dynamics to create an abstraction of the system, the counter example that the decision procedure finds is an \textit{abstract counter example}, and may be spurious. The abstract counter example may be used to seed an execution of the original, concrete system to generate a true counter example.
The novelty of OVERT lies in finding the tightest overapproximation, and therefore one may expect negligible difference between the abstract counter example and the true counter example.
If a real counter example is found, then the property is shown to \textit{fail}. 

We can see in \cref{fig:time}b that the feasibility approach was able to prove the same number of properties as the explicit reachable set computation approach. 
We would expect that the feasibility approach should be able to prove the examples that can be proven with explicit reachable set computation, as the feasibility approach incurs less overapproximation.
In some problems, the feasibility approach may be able to prove examples that cannot be proven with the explicit reachable set computation approach.

The solver was able to find real counter examples for problems C1, C2, and C3; however, this is trivial due to the fact that the intersection of the reachable sets and goal set was empty, as can be seen in the top right plot of \cref{fig:car_tora}. In other words, every trace from the starting set is a counter example. 

\Cref{fig:time}b also shows the compute time of the feasibility experiments.
The compute times for these experiments are reliably shorter than those of the reachable set computation problems, but generally on the same order of magnitude. The TORA problem is a marked exception, where the feasibility framing of the problem is 1 to 2 orders of magnitude faster for problems T1, T2 and T3. The compute time needed to solve problem T3 was reduced from more than 18 hours to less than 9 minutes. 
We hypothesize that there is some interaction between the wide layers (100 neurons) of the T3 control policy and solving the optimization problems over state variables that are part of reachable set computation.
Solving a feasibility problem involves either finding a single feasible solution or showing that no feasible solution exists for each timestep, whereas solving a reachable set computation problem translates to solving $2n$ optimization problems, where $n$ is the state dimension. 
Given the timing results, if an explicit reachable set is not needed, we recommend solving a feasibility problem instead of explicitly computing the reachable set.

\section{Extensions}
\label{sec:exten}
This section discusses different ways to further improve OVERT, both in terms of finding a tighter reachable set and in terms of compute time. 

\subsection{Input splitting}
Solving reachability problems over large input domains and long time horizons can be computationally expensive. 
A large input domain means that a large region of the nonlinear dynamics function must be overapproximated, possibly resulting in a larger number of linear and nonlinear constraints.
A large input domain may also span a large number of activation patterns in the neural network.
Even if the initial set is small, solving over a long time horizon can lead to the reachable sets growing due to overapproximation error, which introduces the same two problems. 
One possible technique to address this problem is to split the input set, as was demonstrated by ~\citet{wu2020parallelization}. 
The split problems are independent of each other, and therefore, can be run in parallel, potentially reducing the runtime.
Input splitting can also serve to reduce the looseness of the overapproximation of the reachable set~\citep{xiang2018specification}.
The reachable set is then no longer represented as a single hypercube but as the overlap of several smaller hypercubes.
We ran preliminary trials of this strategy and found that the reachable set computed was tighter, and leave a more complete exploration of the effect on runtime to future work. 

\subsection{Effect of $L_1$ regularization}
It is well-known that an $L_1$ regularization term in the loss function, when training neural networks, promotes sparsity of the weight matrices. Fewer non-zero weights can lead to a simpler verification problem, as there will be fewer variables or simpler constraints in the verification query \citep{narodytska2019search}.
We tested this idea here. We retrained a network with a configuration similar to the T1 problem (i.e. three hidden layers of 25 neurons each) with $L_1$-regularizing terms for the weights of the hidden layers.  
Interestingly, we observed a significant speedup for reachable set computation problems. 
While this is an exciting finding, we acknowledge more studies are necessary to further elucidate the effect of $L_1$ regularization. 
In addition, this should be studied in conjunction with the effect of $L_1$ regularization on the performance of the network. 

\subsection{Smooth Activation Functions}
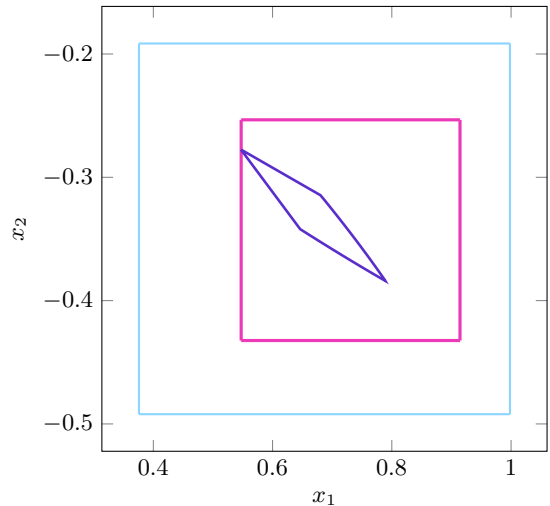
\begin{figure}
    \centering
    \begin{tikzpicture}[]
\begin{axis}[ylabel = {$x_2$}, xlabel = {$x_1$}, width=7.5cm, height=7.5cm]

\addplot+ [solid, concrete_color, thick, mark=none]coordinates {
(0.3759103886443983, -0.1914905244822584)
(0.9981420390905879, -0.1914905244822584)
};
\addplot+ [solid, concrete_color, thick, mark=none]coordinates {
(0.3759103886443983, -0.4921063419393615)
(0.9981420390905879, -0.4921063419393615)
};
\addplot+ [solid, concrete_color, thick, mark=none]coordinates {
(0.3759103886443983, -0.4921063419393615)
(0.3759103886443983, -0.1914905244822584)
};
\addplot+ [solid, concrete_color, thick, mark=none]coordinates {
(0.9981420390905879, -0.4921063419393615)
(0.9981420390905879, -0.1914905244822584)
};

\addplot+ [solid, symbolic_color, very thick, mark=none]coordinates {
(0.5473266075705806, -0.2534008819214195)
(0.9143442720060077, -0.2534008819214195)
};
\addplot+ [solid, symbolic_color, very thick, mark=none]coordinates {
(0.5473266075705806, -0.4323620903309122)
(0.9143442720060077, -0.4323620903309122)
};
\addplot+ [solid, symbolic_color, very thick, mark=none]coordinates {
(0.5473266075705806, -0.4323620903309122)
(0.5473266075705806, -0.2534008819214195)
};
\addplot+ [solid, symbolic_color, very thick, mark=none]coordinates {
(0.9143442720060077, -0.4323620903309122)
(0.9143442720060077, -0.2534008819214195)
};
\addplot+ [solid, mc_color, line width=1pt, mark=none]coordinates {
(0.5471997256747485, -0.27767713174398945)
(0.6463566435286959, -0.3420581143221929)
(0.6478884139575589, -0.3425373344290319)
(0.6482466651385007, -0.3426482049632922)
(0.6627727362709793, -0.3470939781569834)
(0.6729276377170135, -0.3501712386785986)
(0.6852756794284404, -0.35389931315621304)
(0.7009465622695698, -0.3585837317687943)
(0.7024325761251602, -0.3590256147177802)
(0.715044760044155, -0.3627564138471166)
(0.7207846599396871, -0.36444501766721)
(0.7257114332785825, -0.36588649179807736)
(0.7307085385693928, -0.3673427580168962)
(0.7353994368775469, -0.36870914324926213)
(0.7415709966986885, -0.3704984996422724)
(0.7424260562384706, -0.37074606654656955)
(0.7505651378174221, -0.3730939428116312)
(0.7507149442635914, -0.3731370741070476)
(0.7566901792774179, -0.3748490471649801)
(0.7668080990261059, -0.3777231140830735)
(0.7750420232588094, -0.3800586042591248)
(0.7809258804737793, -0.3817210174320962)
(0.7890884642016511, -0.383984969763314)
(0.788664730696202, -0.3836185423323215)
(0.7880381835577099, -0.38310355867794926)
(0.7842142012494407, -0.38046330159269115)
(0.777491785276938, -0.3758514010280822)
(0.7750028272183683, -0.3741564471516523)
(0.7695686078667147, -0.3704807739734642)
(0.7678130942677244, -0.369302210165477)
(0.7576977607673032, -0.36258265066961426)
(0.7489563491128745, -0.3568780855176096)
(0.7436819425670478, -0.3534668621595706)
(0.7396238817497977, -0.35086766766551564)
(0.7346964245661594, -0.34772499438089166)
(0.7303174569650515, -0.3449493075904775)
(0.7241646718028057, -0.34110022136426943)
(0.7201124963230537, -0.3385725446876144)
(0.7132725109914491, -0.33434015430301506)
(0.7045994248012679, -0.3290270286925469)
(0.6994773459045001, -0.3259190608262985)
(0.6923041064813954, -0.32160679018206617)
(0.6859834356934774, -0.3178415300461714)
(0.6828954509286358, -0.31601626841056324)
(0.6822728683306115, -0.3156590825975257)
(0.6803863908690018, -0.3145875528439766)
(0.679129431858195, -0.314211366978287)
(0.5474931956399735, -0.27766013117164545)
};
\end{axis}

\end{tikzpicture}
    \caption{Reachable set computation problem using single pendulum dynamics with a neural network control policy that has only one hidden layer with 25 neurons with $\tanh$ activation functions.  Boxes indicate reachable sets at timestep $n_t=20$ obtained via 1-step concretization (blue) and hybrid-symbolic computation (pink). The convex hull of Monte Carlo simulations is shown in purple. 
    \label{fig:tanh}}
\end{figure}

Earlier, we described how OVERT has been designed for networks with piecewise linear activation functions such as ReLUs.
It is, however, possible to use OVERT with smooth activation functions such as $\tanh$ or sigmoid by overapproximating each smooth nonlinear activation function in the same manner that the system dynamics are overapproximated. 
We do not expect such an approach to scale as well as problems containing ReLU networks, but ran an initial exploration.
We trained a new network for the inverted pendulum problem with $\tanh$ activation functions. 
The network has a single layer with 25 neurons, which is purposefully smaller than those studied earlier. 
Reachable sets were computed for $n=20$ timesteps, with symbolic queries performed at timesteps 10 and 20.
The concrete and symbolic reachable sets produced approximate the convex hull of Monte Carlo simulations closely, as can be seen in \cref{fig:tanh}.
Although every smooth nonlinearity is approximated, the accuracy provided by OVERT's optimal piecewise-linear overapproximations still yields tight reachable sets. 

\section{Conclusion}
In this paper, we introduced a methodology for analyzing closed-loop discrete-time dynamical systems with neural network control policies from a safety perspective. 
Such systems are typically governed by nonlinear dynamical updates, which limit the use of existing neural verification tools. 
To handle this, we overapproximate the nonlinear dynamics with piecewise linear relations. The piecewise linear relations are then encoded into an MIP and solved using a commercial MIP solver. 
Our experiments on a proof of concept implementation demonstrate that OVERT outperforms naive adaptations of existing tools for use with nonlinear discrete-time closed-loop systems with neural network control policies in both tightness of computed reachable sets and speed of computation. 
Our work contributes to ongoing efforts to apply verification tools to real-world cyber physical systems, such as autonomous vehicles and airplane collision avoidance systems.  

\acks{The authors would like to thank Clark Barrett, Christopher Strong, Sydney Katz, Aleksandar Zeljic, Kyle Julian, Changliu Liu, Tomer Arnon, Joe Vincent, and Ransalu Senanayake for their support and feedback. This work was supported by AFRL and DARPA under contract FA8750-18-C-0099. A. Maleki acknowledges the support of the Natural Sciences and Engineering Research Council of Canada (NSERC). }

\appendix
\section{Algorithms for Overapproximation} \label{sec:app1}
This section provides detailed derivations of our algorithms for finding a closed form piecewise-linear overapproximation of any function mapping $\mathbb R^n \rightarrow \mathbb R$. 
\Cref{alg:overtstepone} breaks the problem down into 
finding optimal piecewise-linear overapproximations for one-dimensional functions mapping $\mathbb R \rightarrow \mathbb R$ (\cref{app:overapprox1D}).
Each piecewise-linear overapproximation is formed using an upper and lower bound as is specified in \cref{alg:overtsteptwo}.
Finally, each piecewise-linear bound is represented as a single closed-form function (\cref{sec:closed-form}).

\subsection{Optimality of the Midpoint Tangent Bound}
\label{sec:app0}
Consider a function $f$ that is strictly concave over the interval $[x_{i-1}, x_i]$. Over this interval, we seek to construct the tightest upper bound for this function that consists of a single tangent line segment. The upper bound $g_i(x,\alpha)$ is defined to be tangent to $f$ at some point $\alpha$, $\alpha \in [x_{i-1}, x_i]$:

\begin{equation*}
    g_i(x, \alpha) = f'(\alpha)\cdot (x-\alpha) + f(\alpha)
\end{equation*}

The ``tightest" upper bound is defined as the bound which minimizes the area between the bound and the function:
\begin{align*}
    \min_{\alpha} \int_{x_{i-1}}^{x_i} \left ( g_i(x, \alpha) - f(x) \right ) \mathrm{d} x 
\end{align*}

As the area under the function $f$ does not depend on $\alpha$, we focus on minimizing the area $A(\alpha)$ under the bound:
\begin{equation*}
    A(\alpha) = \int_{x_{i-1}}^{x_i} g_i(x, \alpha) \mathrm{d} x 
\end{equation*}

The area $A(\alpha)$ under the bound can alternately be expressed using the midpoint integral approximation, which is exact for a linear function:

\begin{align*}
    A(\alpha) &= g(x_{mid}, \alpha)\cdot (x_i - x_{i-1})
\end{align*}
where $\alpha$ is the tangent point and $x_{mid}$ is the midpoint: $x_{mid} = \frac{x_i + x_{i-1}}{2}$ (dropping the subscript $i$ for simplicity).
As the term $(x_i - x_{i-1})$ is fixed, we can further focus on minimizing only $g(x_{mid}, \alpha)$.

Consider choosing $\alpha = x_{mid}$. The term $g(x_{mid}, \alpha)$ evaluates to:
\begin{align*}
    g_i(x_{mid}, \alpha = x_{mid}) &= f'(x_{mid})\cdot (x_{mid}-x_{mid}) + f(x_{mid}) \\
    &= f(x_{mid})
\end{align*}
Or, in other words, $g_i(x_{mid}, \alpha = x_{mid})$ lies \textit{on} the function. 

Next, consider choosing $\alpha = \beta$, where $\beta \neq x_{mid}$ is any other point in the interval $[x_i - x_{i-1}]$ different from $x_{mid}$. While the bound $g_i(x, \beta)$ lies on the function at $x=\beta$:
\begin{align*}
    g_i(x=\beta, \beta) &= f'(\beta)\cdot (\beta-\beta) + f(\beta) \\
     &= f(\beta)
\end{align*}

evaluating $g_i(x, \beta)$  at any other point $x$ along the bound, including $x_{mid}$, necessarily produces points \textit{above} the function:
\begin{equation*}
    g_i(x, \beta) > f(x)~, ~ \forall x \neq \beta~,~ x\in [x_i - x_{i-1}]
\end{equation*}
as the tangent to a strictly concave function touches the function at exactly one point and lies above the function at all other points. 
Consequently, 
\begin{equation*}
    g_i(x_{mid}, \beta) > f(x_{mid})
\end{equation*}
and because $g_i(x_{mid}, \alpha = x_{mid}) = f(x_{mid})$ this implies that 
\begin{equation*}
     g_i(x_{mid}, \beta) > g_i(x_{mid}, x_{mid})
\end{equation*}
and therefore that $\alpha = x_{mid}$ is the minimizer of $ g_i(x_{mid}, \alpha)$ over the interval $[x_i - x_{i-1}]$. This further implies, as reasoned above, that 
\begin{align*}
    x_{mid} &= \arg\min_{\alpha} A(\alpha) \\
    &= \arg\min_{\alpha} \int_{x_{i-1}}^{x_i} \left ( g_i(x, \alpha) - f(x) \right ) \mathrm{d} x 
\end{align*}
or, in other words, that the tangent line at the midpoint is the tightest tangent bound for a strictly concave function over a fixed interval. 

\begin{figure}[h!]
\centering
\begin{tikzpicture}[]
\begin{axis}[
  legend pos = {south east},
  xlabel = {$x$}
]

\addplot+[
  red, mark options={fill=red}
] coordinates {
  (0.5, 2.75)
};
\addlegendentry{{}{$g_i(x_{mid}, x_{mid})$}}

\addplot+[
  red, mark options={fill=red}
] coordinates {
  (0.5, 3.3125)
};
\addlegendentry{{}{$g_i(x_{mid}, \beta) $}}

\addplot+[
  blue, mark=none
] coordinates {
  (-1.0, -1.0)
  (-0.95, -0.8024999999999998)
  (-0.9, -0.6099999999999999)
  (-0.85, -0.4225000000000003)
  (-0.8, -0.2400000000000002)
  (-0.75, -0.0625)
  (-0.7, 0.11000000000000032)
  (-0.65, 0.2775000000000003)
  (-0.6, 0.4399999999999995)
  (-0.55, 0.5974999999999997)
  (-0.5, 0.75)
  (-0.45, 0.8975)
  (-0.4, 1.0400000000000003)
  (-0.35, 1.1774999999999998)
  (-0.3, 1.3099999999999998)
  (-0.25, 1.4375)
  (-0.2, 1.56)
  (-0.15, 1.6775000000000002)
  (-0.1, 1.7899999999999998)
  (-0.05, 1.8975)
  (0.0, 2.0)
  (0.05, 2.0975)
  (0.1, 2.19)
  (0.15, 2.2775)
  (0.2, 2.36)
  (0.25, 2.4375)
  (0.3, 2.5100000000000002)
  (0.35, 2.5775)
  (0.4, 2.64)
  (0.45, 2.6975)
  (0.5, 2.75)
  (0.55, 2.7975)
  (0.6, 2.84)
  (0.65, 2.8775)
  (0.7, 2.91)
  (0.75, 2.9375)
  (0.8, 2.96)
  (0.85, 2.9775)
  (0.9, 2.99)
  (0.95, 2.9975)
  (1.0, 3.0)
  (1.05, 2.9975)
  (1.1, 2.9899999999999998)
  (1.15, 2.9775)
  (1.2, 2.96)
  (1.25, 2.9375)
  (1.3, 2.91)
  (1.35, 2.8775)
  (1.4, 2.84)
  (1.45, 2.7975)
  (1.5, 2.75)
  (1.55, 2.6975)
  (1.6, 2.6399999999999997)
  (1.65, 2.5775)
  (1.7, 2.5100000000000002)
  (1.75, 2.4375)
  (1.8, 2.36)
  (1.85, 2.2775)
  (1.9, 2.1900000000000004)
  (1.95, 2.0975)
  (2.0, 2.0)
};
\addlegendentry{{}{$f(x)$}}

\addplot+[
  solid, red, mark=none
] coordinates {
  (-1.0, 1.25)
  (-0.95, 1.3)
  (-0.9, 1.35)
  (-0.85, 1.4)
  (-0.8, 1.45)
  (-0.75, 1.5)
  (-0.7, 1.55)
  (-0.65, 1.6)
  (-0.6, 1.65)
  (-0.55, 1.7)
  (-0.5, 1.75)
  (-0.45, 1.8)
  (-0.4, 1.85)
  (-0.35, 1.9)
  (-0.3, 1.95)
  (-0.25, 2.0)
  (-0.2, 2.05)
  (-0.15, 2.1)
  (-0.1, 2.15)
  (-0.05, 2.2)
  (0.0, 2.25)
  (0.05, 2.3)
  (0.1, 2.35)
  (0.15, 2.4)
  (0.2, 2.45)
  (0.25, 2.5)
  (0.3, 2.55)
  (0.35, 2.6)
  (0.4, 2.65)
  (0.45, 2.7)
  (0.5, 2.75)
  (0.55, 2.8)
  (0.6, 2.85)
  (0.65, 2.9)
  (0.7, 2.95)
  (0.75, 3.0)
  (0.8, 3.05)
  (0.85, 3.1)
  (0.9, 3.15)
  (0.95, 3.2)
  (1.0, 3.25)
  (1.05, 3.3)
  (1.1, 3.35)
  (1.15, 3.4)
  (1.2, 3.45)
  (1.25, 3.5)
  (1.3, 3.55)
  (1.35, 3.6)
  (1.4, 3.65)
  (1.45, 3.7)
  (1.5, 3.75)
  (1.55, 3.8)
  (1.6, 3.85)
  (1.65, 3.9)
  (1.7, 3.95)
  (1.75, 4.0)
  (1.8, 4.05)
  (1.85, 4.1)
  (1.9, 4.15)
  (1.95, 4.2)
  (2.0, 4.25)
};

\addplot+[
  dashed, red, mark=none
] coordinates {
  (-1.0, -0.4375)
  (-0.95, -0.3125)
  (-0.9, -0.1875)
  (-0.85, -0.0625)
  (-0.8, 0.0625)
  (-0.75, 0.1875)
  (-0.7, 0.3125)
  (-0.65, 0.4375)
  (-0.6, 0.5625)
  (-0.55, 0.6874999999999999)
  (-0.5, 0.8125)
  (-0.45, 0.9375)
  (-0.4, 1.0625)
  (-0.35, 1.1875)
  (-0.3, 1.3125)
  (-0.25, 1.4375)
  (-0.2, 1.5625)
  (-0.15, 1.6875)
  (-0.1, 1.8125)
  (-0.05, 1.9375)
  (0.0, 2.0625)
  (0.05, 2.1875)
  (0.1, 2.3125)
  (0.15, 2.4375)
  (0.2, 2.5625)
  (0.25, 2.6875)
  (0.3, 2.8125)
  (0.35, 2.9375)
  (0.4, 3.0625)
  (0.45, 3.1875)
  (0.5, 3.3125)
  (0.55, 3.4375)
  (0.6, 3.5625)
  (0.65, 3.6875)
  (0.7, 3.8125)
  (0.75, 3.9375)
  (0.8, 4.0625)
  (0.85, 4.1875)
  (0.9, 4.3125)
  (0.95, 4.4375)
  (1.0, 4.5625)
  (1.05, 4.6875)
  (1.1, 4.8125)
  (1.15, 4.9375)
  (1.2, 5.0625)
  (1.25, 5.1875)
  (1.3, 5.3125)
  (1.35, 5.4375)
  (1.4, 5.5625)
  (1.45, 5.6875)
  (1.5, 5.8125)
  (1.55, 5.9375)
  (1.6, 6.0625)
  (1.65, 6.1875)
  (1.7, 6.3125)
  (1.75, 6.4375)
  (1.8, 6.5625)
  (1.85, 6.6875)
  (1.9, 6.8125)
  (1.95, 6.9375)
  (2.0, 7.0625)
};

\end{axis}
\end{tikzpicture}
\caption{Graphical depiction of midpoint optimality proof.}
\end{figure}
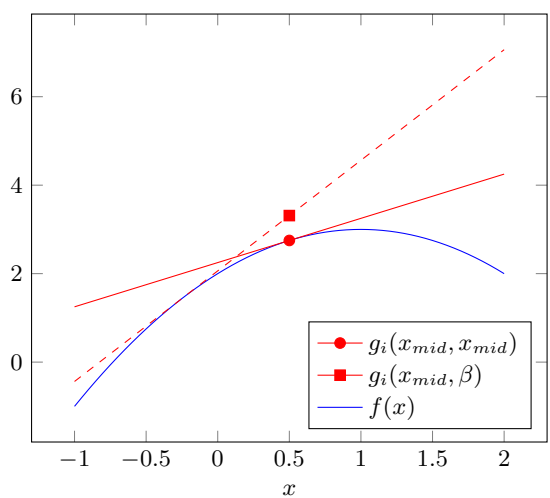

\subsection{Bounding 1D (scalar) functions} \label{app:overapprox1D}
Consider a scalar function mapping  $\mathbin{f}{:}{~\mathbb{R}}{\rightarrow}{\mathbb{R}}$.  
Our goal is to find the tightest piecewise-linear upper and lower bound functions over domain $[p,q]$. 
In order to find such bounds, we need to divide domain $[p,q]$ into intervals within which $f''(x)$ does not change sign. That is, within each interval $[a,b]$, $f(x)$ is either convex or concave. 
In order to have a continuous function and express the piecewise-linear function in closed form, as described in section \cref{sec:closed-form}, the bound from each convex or concave interval must be coincident to the function at the interval endpoints.

More formally, given a natural number $n$, our goal is to find the tightest piecewise-linear upper or lower bound function $g(x)$ for $f$ over interval $[a,b]$ that is composed of $n$ linear pieces (see \cref{fig:fig_g_convex}). More specifically, we choose $n-1$ points $x_1,x_2, \cdots, x_{n-1}$ within the interval $[a,b]$ such that: $a=x_0 < x_1 < x_2 < \cdots < x_{n-1} < x_n=b$. Then, 
\begin{equation}\label{eq:g(x)}
g_i(x) = \frac{y_i - y_{i-1}}{x_i - x_{i-1}} \left( x-x_i\right) + y_i, ~~~ x_{i-1} \leq x \leq x_i
\end{equation}
where points $(x_i,y_i), i=1,2,\cdots n-1$ need to be specified, and $g(x_0=a)=f(a)$ and $g(x_n=b)=f(b)$.  The following algorithms then specify how to \emph{optimally} find points $(x_i,y_i)$ in \cref{eq:g(x)} within each interval.

\subsubsection{Upper (lower) bound for a 1D convex (concave) function} \label{sec:app-conv}

The tightest upper bound function $g(x)$ in \cref{eq:g(x)} for a convex function $f(x)$ over interval $[a,b]$ is specified by selecting points $x_i$ that satisfy the following system of equations:
\begin{align}\label{eq:optimility_convex}
   f'(x_i) = \frac{f(x_{i+1})-f(x_{i-1})}{x_{i+1}-x_{i-1}},~~ \text{for}~~ i=1,2,\cdots n-1, \\
   y_i = f(x_i) 
\end{align}
To ensure continuity $x_0=a, x_n=b$, and $y_0=f(x_0), y_n=f(x_n)$.   

\paragraph{Proof:} Here we prove that $g(x)$ is a tight upper bound for a convex function. One can similarly show that this is a tight lower bound for a concave function. 

Consider function $f(x)$ and its overapproximation $g(x; \mathbin{x_0}{:}{x_n})$, where $\mathbin{x_0}{:}{x_n}$ denotes $n+1$ points $x_0, x_1, x_2, \cdots, x_n$ along the $x$-axis.
Function $g$ is comprised of $n$ linear segments $g_i(x)$ between $x_{i-1}$ and $x_i$ as shown in \cref{fig:fig_g_convex}. 
Since $f(x)$ is convex, application of Jensen's inequality straightforwardly yields $g_i (x) \geq f(x)$ for $ x_{i-1}\leq x \leq x_i$, suggesting that $g(x) \geq f(x), \forall x \in [a,b]$.  
Finding the tightest possible $g$ amounts to minimizing the shaded area in \cref{fig:fig_g_convex}, or equivalently, solving the following minimization problem:
\begin{equation}\label{eq:min_object}
\min_{x_1:x_{n-1}} \int_a^b \left[g(x; \mathbin{x_0}{:}{x_n}) - f(x) \right] \mathrm{d}x
\end{equation}
Notice that $f(x)$ is a constant with respect to $\mathbin{x_0}{:}{x_n}$, and can be removed from the objective. 

We rewrite \cref{eq:min_object}:
\begin{align*}
\min_{x_1:x_{n-1}} \int_a^b \left[g(x; \mathbin{x_0}{:}{x_n}) - f(x) \right] \mathrm{d}x = &
 \min_{x_1:x_{n-1}} \sum_{i=0}^{n-1} \int_{x_{i-1}}^{x_{i}} g_i(x; x_{i-1}, x_{i}) \mathrm{d}x  
 \\ = & 
 \min_{x_1:x_{n-1}} \sum_{i=0}^{n-1} \int_{x_{i-1}}^{x_{i}} \left[\frac{f(x_{i}) - f(x_{i-1})}{x_{i} - x_{i-1}} (x-x_{i-1}) + f(x_{i-1}) \right] \mathrm{d}x
 \end{align*}
 Using the fact that the midpoint approximation of the integral of a line is exact, we can write the expression as:
 \begin{align*}
  = & 
  \min_{x_1:x_{n-1}} \sum_{i=0}^{n-1} \left( \frac{f(x_{i}) + f(x_{i-1})}{2}\left(x_{i} - x_{i-1} \right)\right) 
\end{align*}
We take derivative of the objective with respect to $x_j; j=1,2,\cdots, n-1$  and set to zero:
\begin{align*}
0 = &\frac{\partial}{\partial x_j} \sum_{i=0}^{n-1} \left( \frac{f(x_{i}) + f(x_{i-1})}{2}\left(x_{i} - x_{i-1} \right)\right) 
\\
=& \frac{\partial}{\partial x_j} \left( \frac{f(x_{j}) + f(x_{j-1})}{2}\left(x_{j} - x_{j-1} \right)\right) 
 + \frac{\partial}{\partial x_j} \left( \frac{f(x_{j}) + f(x_{j+1})}{2}\left(x_{j+1} - x_{j} \right)\right) 
 \\
 =& \frac{1}{2} \left[ f'(x_j)(x_j-x_{j-1}) + f(x_j) + f(x_{j-1}) + f'(x_j)(x_{j+1}-x_{j}) - f(x_j) - f(x_{j+1}) \right]
 \\
 =& \frac{1}{2} \left[ f'(x_j)(x_{j+1} -x_{j-1}) +  f(x_{j-1}) -f(x_{j+1}) \right]
\end{align*}
which simplifies to
\begin{align} \label{eq:convex_optim_repeat}
f'(x_j) = \frac{f(x_{j+1}) -f(x_{j-1})}{x_{j+1} -x_{j-1}}\BlackBox
\end{align}
While we do not show here that this stationary point is a minimum, empirically, we have observed that it appears to be.
Points $(x_i,y_i)$ satisfying the optimality condition in \cref{eq:convex_optim_repeat} are found using a numerical routine provided by NLsolve. If there is error in the numerical routine and the $x_i$ are not optimally placed, the bound produced will still be valid and continuous, as any secant connecting two points $(x_{i-1}, f(x_{i-1}))$ and $(x_i,f(x_i))$ of a convex function is a valid upper bound over the interval $[x_{i-1}, x_i]$.

\subsubsection{Upper (lower) bound for a 1D concave (convex) function}
Consider a function $\mathbin{f}{:}{\mathbb{R}\rightarrow \mathbb{R}}$ that is concave over interval $[a,b]$. Similar to the derivation procedure in \cref{sec:app-conv}, we find the optimum points $x_i$ in \cref{eq:g(x)} by minimizing the area between $g(x)$ and $f(x)$. 

The tightest upper bound function $g(x)$ for a concave function $f(x)$ over interval $[a,b]$ is specified by points $x_i, i=1,2, \cdots, n-1$ that satisfy:
\begin{equation} \label{eq:optimility_concave}
x_i = h \left(\frac{x_{i-1}+x_i}{2}, \frac{x_i + x_{i+1}}{2} \right)
\end{equation}
where 
\begin{equation} \label{eq:h_app}
h(\alpha, \beta) = \frac{\beta f'(\beta)-\alpha f'(\alpha)}{f'(\beta)-f'(\alpha)} - \frac{f(\beta)-f(\alpha)}{f'(\beta)-f'(\alpha)}
\end{equation}
The $y_i$ values are given by
\begin{align*}
y_i &= g_i(x_i) \nonumber \\
&= f'\left(\frac{x_{i-1}+x_i}{2}\right) \left(x_i - \frac{x_i+x_{i-1}}{2}\right) + f\left(\frac{x_i+x_{i-1}}{2} \right) \nonumber \\
&= f'\left(\frac{x_{i-1}+x_i}{2}\right) \left(\frac{x_i-x_{i-1}}{2}\right) + f\left(\frac{x_i+x_{i-1}}{2} \right)
\end{align*}

\paragraph{Proof:}
Consider line segment $g_i(x)$ that is an upper bound for the concave function $f(x)$ over the interval $[x_{i-1},x_i]$. Since $f(x)$ is concave, $g_i(x)$ may not intersect $f(x)$ at more than one points; otherwise, between the intersecting points, by Jensen's inequality, $g_i(x) <  f(x)$. That means, $g_i(x)$ can at best be tangent to $f(x)$. Denote the tangent point $\alpha \in [x_{i-1},x_i]$. We showed in \cref{sec:app0} that if \[\alpha= \frac{x_{i-1} + x_i}{2}\] the area between $f(x)$ and $g_i(x)$ is minimized. That means once we find $x_{i-1}$ and $x_i$, function $g_i$ is the line that is tangent to $f(x)$ at  $(x_{i-1} + x_i)/2$. We now turn our attention to optimally distribute the $x_i$'s. Similar to the previous algorithm, we would like to minimize the area between $f(x)$ and $g(x)$:
\begin{align*}
\min_{x_1:x_{n-1}} \int_a^b \left[g(x; \mathbin{x_0}{:}{x_n}) - f(x) \right] \mathrm{d}x = &
 \min_{x_1:x_{n-1}} \sum_{i=1}^{n} \int_{x_{i-1}}^{x_{i}} g_i(x; x_{i-1}, x_{i}) \mathrm{d}x  
 \\ = & 
 \min_{x_1:x_{n-1}} \sum_{i=1}^{n} \int_{x_{i-1}}^{x_{i}} f'\left(\frac{x_{i-1}+x_i}{2}\right) (x-\frac{x_{i-1}+x_i}{2}) + f\left(\frac{x_{i-1}+x_i}{2}\right) \mathrm{d}x
  \\ = & 
  \min_{x_1:x_{n-1}} \sum_{i=1}^{n} f\left(\frac{x_{i-1}+x_i}{2}\right) \left(x_i- x_{i-1}\right) 
\end{align*}
Taking derivative with respect to $x_j, j=1,2,\cdots ,n-1$ and setting to zero:
\begin{align*}
0 = &\frac{\partial}{\partial x_j} \sum_{i=1}^{n} f\left(\frac{x_{i-1}+x_i}{2}\right) \left(x_i- x_{i-1}\right) 
\\
=& \frac{\partial}{\partial x_j} \left( f\left(\frac{x_{j-1}+x_j}{2}\right) \left(x_j- x_{j-1}\right)  \right) 
 + \frac{\partial}{\partial x_j} \left( f\left(\frac{x_{j+1}+x_j}{2}\right) \left(x_{j+1}- x_{j}\right)  \right) 
 \\
 =& \frac{1}{2} f'\left(\frac{x_{j-1}+x_{j}}{2}\right) \left(x_{j}- x_{j-1}\right) + f\left(\frac{x_{j-1}+x_j}{2}\right) + 
    \frac{1}{2} f'\left(\frac{x_{j+1}+x_j}{2}\right) \left(x_{j+1}- x_{j}\right) - f\left(\frac{x_{j+1}+x_j}{2}\right)
\end{align*}
which simplifies to
\begin{equation}\label{eq:alg1}
\begin{split}
	\frac{1}{2} f'\left(\frac{x_i+x_{i-1}}{2} \right) \left(x_i-x_{i-1}\right) + f\left(\frac{x_i+x_{i-1}}{2} \right) = \\ 
	\frac{1}{2} f'\left(\frac{x_i+x_{i+1}}{2} \right) \left(x_i-x_{i+1}\right) + f\left(\frac{x_i+x_{i+1}}{2} \right) 
\end{split}
\end{equation}  
Re-arranging \cref{eq:alg1} and solving for $x_i$, we can reproduce \cref{eq:optimility_concave}. It is worth showing that the optimality constraint (\cref{eq:alg1}) also enforces continuity over points $x_1, x_2, \ldots, x_{n-1}$. To show this, we re-arrange \cref{eq:alg1}:
\[
	f'\left(\frac{x_i+x_{i-1}}{2} \right) \left(\frac{x_i-x_{i-1}}{2}\right) + f\left(\frac{x_i+x_{i-1}}{2} \right) = 
	f'\left(\frac{x_i+x_{i+1}}{2} \right) \left(\frac{x_i-x_{i+1}}{2}\right) + f\left(\frac{x_i+x_{i+1}}{2} \right) 
\]
\[\Rightarrow\]
\[ f'\left(\frac{x_i+x_{i-1}}{2} \right) \left(x_i - \frac{x_i+x_{i-1}}{2}\right) + f\left(\frac{x_i+x_{i-1}}{2} \right) = 
	f'\left(\frac{x_i+x_{i+1}}{2} \right) \left(x_i - \frac{x_i+x_{i+1}}{2}\right) + f\left(\frac{x_i+x_{i+1}}{2} \right)\]
\[ \Rightarrow\]
\[ g_{i}(x_i) = g_{i+1}(x_i)\]
which indicates that $g(x)$ is continuous within interval $(a,b)$. 
\BlackBox

While we do not show here that the stationary point satisfying \cref{eq:optimility_concave} is a minimum, empirically we have observed that the stationary point is a unique minimum. Proof for arbitrary concave $f$ is still open.

If function $f(x)$ contains both convex and concave sub-intervals, the application of \cref{eq:optimility_convex} and \cref{eq:optimility_concave} leads to discontinuity at the inflection points. Note that continuity is necessary in order to be able to express piecewise-linear function in closed form; see \cref{sec:closed-form}. Discontinuity results because the value of over-approximating function $g(x)$ at the endpoints of concave sub-intervals is not equal to that of $f(x)$. To address this, we enforce that the bound $g(x)$ is tangent to the function $f$ at both $a$ and $b$, which implies that $g(a) = f(a) $ and $g(b)=f(b)$ on all concave sub-intervals (notice that this condition is already satisfied on convex sub-intervals). Therefore, the bound is given by points $(x_i, y_i)$ that satisfy:
\begin{subequations}
\begin{align*}
x_0 =& a \\
x_1 =& h\left(a, \frac{x_1 + x_{2}}{2} \right)\\
x_i =& h \left(\frac{x_{i-1}+x_i}{2}, \frac{x_i + x_{i+1}}{2} \right),~~~ i=2, \cdots n-2\\
x_{n-1} =& h \left(\frac{x_{n-1}+x_n}{2}, b \right) \\
x_{n} =& b
\end{align*}
\end{subequations}
where $h$ is given by \cref{eq:h_app}, and $y_i$ is given by:
\begin{subequations}
\begin{align*}
    y_0 =& f(a) \\
    y_1 =& f'(a)(x_1 - a) + f(a) \\
    y_i =& f'\left(\frac{x_{i-1} + x_i}{2}\right)\left(\frac{x_i - x_{i-1}}{2}\right) + f\left(\frac{x_{i-1} + x_i}{2}\right), ~i=2,...,(n-2) \\
    y_{n-1} =& f'(b)(x_{n-1} - b) + f(b) \\
    y_n =& f(b)
\end{align*}
\end{subequations}

Satisfying pairs $(x_i, y_i)$ are again found using a numerical routine provided by NLsolve. If the optimality conditions are not met, the bound will be sound but not continuous. The bound may be repaired by taking $y_i = \max(g_i(x_i), g_{i+1}(x_i))$ if it is an upper bound for a concave function. The resulting line segment forming an upper bound may only be shifted upward by this process and thus will still constitute a valid upper bound. For a lower bound for a convex function, the bound may be analogously repaired by taking $y_i = \min(g_i(x_i), g_{i+1}(x_i))$.

\printbibliography

\end{document}